\documentclass{article}
\usepackage{iclr2027_conference,times}

\usepackage{amsmath,amsfonts,bm}

\def\eqref#1{equation~\ref{#1}}

\def\1{\bm{1}}

\DeclareMathAlphabet{\mathsfit}{\encodingdefault}{\sfdefault}{m}{sl}
\SetMathAlphabet{\mathsfit}{bold}{\encodingdefault}{\sfdefault}{bx}{n}

\usepackage{hyperref}
\usepackage{cleveref}
\usepackage{url}
\usepackage{booktabs}
\usepackage{multirow}
\usepackage{makecell}
\usepackage{array}
\usepackage{tabularx}
\usepackage{ragged2e}
\usepackage{graphicx}
\usepackage{amsmath,amssymb}
\usepackage{enumitem}
\usepackage{xcolor}
\usepackage{pifont}
\definecolor{revisionpink}{RGB}{0,0,0}
\definecolor{newpink}{RGB}{0,0,0}
\definecolor{newblue}{RGB}{0,0,0}
\newcommand{\np}[1]{\textcolor{newpink}{#1}}
\definecolor{compressgreen}{RGB}{0,128,80}
\usepackage{placeins}
\usepackage{algorithm}
\usepackage{caption}
\usepackage{algpseudocode}
\usepackage{wrapfig}
\usepackage{fontawesome5}
\definecolor{Red}{rgb}{0.6,0,0}
\definecolor{Blue}{rgb}{0,0,0.8}
\definecolor{Green}{rgb}{0,0.6,0.9}
\definecolor{airforceblue}{rgb}{0.36, 0.54, 0.66}
\definecolor{ao(english)}{rgb}{0.0, 0.5, 0.0}
\definecolor{azure(colorwheel)}{rgb}{0.0, 0.5, 1.0}
\definecolor{crimson}{rgb}{0.86, 0.08, 0.24}
\definecolor{darkcerulean}{rgb}{0.03, 0.27, 0.49}
\definecolor{cobalt}{rgb}{0.0, 0.28, 0.67}
\definecolor{rosegold}{rgb}{0.72, 0.43, 0.47}
\definecolor{orange-red}{rgb}{1.0, 0.27, 0.0}
\definecolor{mountainmeadow}{rgb}{0.19, 0.73, 0.56}
\definecolor{malachite}{rgb}{0.04, 0.85, 0.32}
\definecolor{darkblue}{rgb}{0.0, 0.0, 0.55}
\definecolor{customred}{rgb}{1, 0.85, 0.85}
\definecolor{customcitecolor}{rgb}{0.7, 0.5, 1}
\definecolor{custompink}{rgb}{0.8, 0.3, 0.3}
\definecolor{customgreen}{rgb}{0.3, 0.8, 0.3}

\Crefname{section}{Section}{Sections}
\Crefname{table}{Table}{Tables}
\crefname{section}{Sec.}{Secs.}
\crefname{table}{Tab.}{Tabs.}
\crefname{figure}{Fig.}{Figs.}
\crefname{appendix}{Sec.}{Secs.}
\crefname{equation}{Eq.}{Eqs.}
\crefname{algorithm}{Algo.}{Algos.}

\hypersetup{
  colorlinks=true,
  citecolor=customcitecolor,
  linkcolor=blue,
  urlcolor=blue
}
\usepackage[normalem]{ulem}

\newcommand{\methodname}[1]{Agent Sociodrama}
\definecolor{rescale}{RGB}{0,128,0}
\usepackage{graphicx}

\title{
  \methodname{}:
  Social-Dynamics-Guided 
  Persona Recasting for LLM Agent Teams
}
\author{Yirui Qi\textsuperscript{1,2}\quad\quad Xiaoming Zhang\textsuperscript{1\dag}\quad\quad Jaehong Yoon\textsuperscript{2\dag}\\
\textsuperscript{1}Beihang University \quad\quad \textsuperscript{2}Nanyang Technological University, Singapore\\
\texttt{\{qyr888,yolixs\}@buaa.edu.cn} \quad\quad \texttt{jaehong.yoon@ntu.edu.sg}
\vspace{0.2in}\\
\parbox{\textwidth}{\faGlobe\ Project Page:~
\href{https://lululu896.github.io/Agent-Sociodrama}{\texttt{https://lululu896.github.io/Agent-Sociodrama}}}}
\iclrfinalcopy

\begin{document}
\maketitle
\renewcommand{\thefootnote}{\fnsymbol{footnote}}
\footnotetext[2]{Corresponding authors.}
\renewcommand{\thefootnote}{\arabic{footnote}}
\lhead{Preprint}

\begin{abstract}
Large language model (LLM) agents are increasingly deployed as teams, yet they fail in characteristically social ways: information holders stay silent, dominant speakers anchor groups on wrong answers, and norm violations go unsanctioned.
We present Agent Sociodrama, a framework for modeling and validating evolving social dynamics in LLM agent teams.
Under fixed personas, agents' emotional states and directed, asymmetric relationships are updated every round by a unified external LLM judge and shape their subsequent interactions.
Across six LLM backbones and five social scenarios, we assess social fidelity through classical social-theory patterns and agreement with human judgments.
The resulting trajectories exhibit structured social patterns and model-specific differences.
We then introduce Trajectory-Guided Persona Recasting (TPR), which diagnoses group-dynamic failures from interaction transcripts and social trajectories and recasts agent personas before rerunning the task.
Across four collaborative benchmarks, TPR improves performance from both heterogeneous and randomly initialized agent persona compositions, yielding up to 83.3\%p of accuracy improvement over the initial casting.
On representative distributed-information cooperation and mixed-motive bargaining tasks, we outperform SOTOPIA and AgentVerse in three of four comparisons, by up to 37.5\%p in Pareto rate.
These results show that validated social trajectories provide actionable signals for repairing LLM agent teams.
\end{abstract}

\section{Introduction}
\label{sec:introduction}

Large language model (LLM) agents are increasingly deployed in teams
\citep{wu2024autogen,hong2024metagpt}: they debate answers
\citep{du2023debate}, pool distributed evidence, teach one another,
negotiate, and manage shared resources. Yet sustained multi-agent
interaction requires more than assigning each agent an initial role.
For a team to function as a social system over time, its members must
form, maintain, and revise relationships in ways that remain coherent
with the interaction. Existing agent systems often encode social
context through static role instructions or unstructured memory,
without an explicit and validated account of how directed relationships
evolve over time.

We take a different route. We argue that longitudinal role play requires treating a team not as a collection of fixed personas, but as a dynamic social system. A persona specifies who an agent is, but not how its relationships with others develop over time, including whom it comes to trust, avoid, align with, or oppose. These relationships must therefore be represented as evolving interaction states. The key challenge is not merely to model changing relations, but to ensure that those changes remain socially faithful over time.

\begin{figure}[t]
    \centering
    \includegraphics[width=\linewidth]{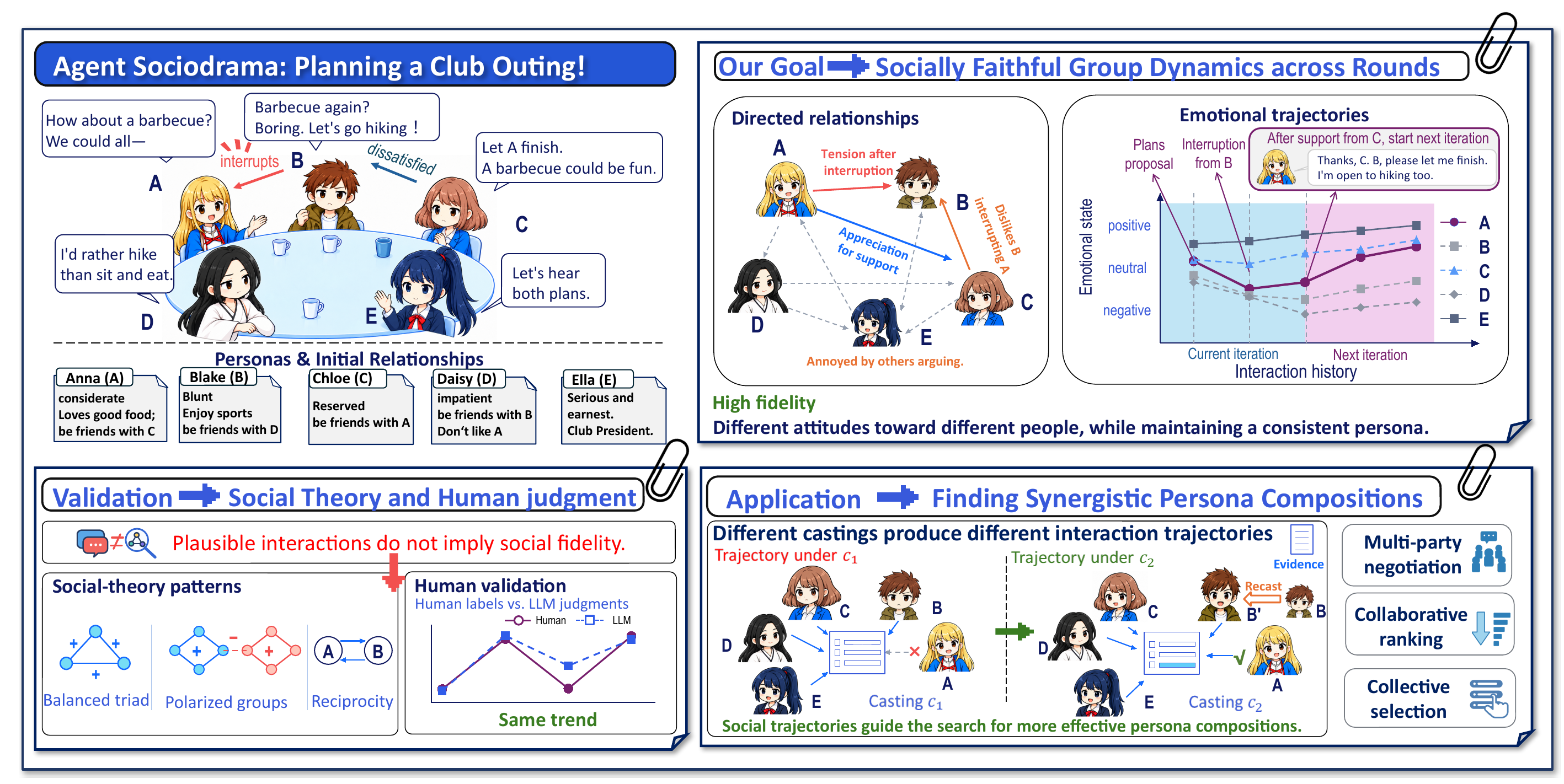}
    \caption{\textbf{Motivation and overview of \methodname{}.}
    \methodname{} models multi-agent interaction through stable personas,
    evolving emotions, and directed relationships.
    We assess the social fidelity of these dynamics against
    social-theory patterns and human judgments, and use social
    trajectories to guide team composition optimization for better
    collaboration.
    }
    \label{fig:concept}
    \vspace{-0.25in}
\end{figure}

To instantiate this idea, we propose \methodname{}, a longitudinal
role-playing framework for LLM agent teams. \methodname{} operationalizes
stable personas together with evolving emotional and relational states,
providing a structured account of how a team's social configuration
develops over repeated interaction.
These dynamics also matter for task-oriented teams. A growing body of
evidence suggests that LLM teams often fail in ways that resemble
classic small-group pathologies rather than limits of individual model
capability: groups that collectively possess the information required
for a correct decision can still converge on an incorrect one
\citep{teams2026experts,li2026hiddenbench}, while individually capable
agents can collectively deplete a shared resource until cooperation
collapses \citep{piatti2024govsim}. These failures have a distinctly
social character. Over repeated interaction, overlooked members can
withdraw, early deference to dominant voices can harden into consensus,
and unchallenged violations can progressively erode cooperation. Yet
dominant approaches to improving multi-agent systems optimize prompts,
topologies, or workflows \citep{zhou2026mass,zhang2025aflow}, treating
the team as a black box and leaving the social mechanisms behind these
failures largely unexamined. Building on \methodname{}'s explicit social
trajectories, we further introduce Trajectory-Guided Persona Recasting
(TPR), a casting controller that diagnoses team composition and
recasts personas to seek more effective task-oriented collaboration.

Our empirical study has two parts. First, we assess the social fidelity of \methodname{} across six LLM drivers and five social scenarios, using a shared external LLM judge with human cross-validation. We test whether the resulting trajectories recover four established regularities of social interaction: structural balance~\citep{heider1946attitudes,cartwright1956structural}, reciprocity~\citep{gouldner1960norm}, conflict-driven polarization~\citep{krackhardt1988informal}, and divergent repair-versus-escalation dynamics~\citep{gottman1994why,pruitt2004social}. These analyses establish \methodname{} as a socially faithful framework for longitudinal multi-agent role play. Second, across four collaborative task families, our controller converts evolving social states into persona recasting decisions that substantially improve team performance. These results demonstrate that socially faithful trajectory modeling can directly inform persona recasting and improve multi-agent collaboration.

Our contributions are:
\vspace{-0.1in}
\begin{itemize}[leftmargin=*,itemsep=0.03in]
  \item \textbf{A dynamic multi-agent role-playing framework for LLM teams.}
  \methodname{} jointly models stable personas, evolving emotional states,   and directed, asymmetric relational states, allowing multi-agent   interaction to unfold as a sustained social process.

  \item \textbf{Cross-model validation of social fidelity.}
  Across six LLMs and five social scenarios, we validate
  \methodname{} against social theory and human evaluation, showing that its long-horizon relational dynamics are structured, interpretable,  and socially coherent.

  \item \textbf{Trajectory-guided team composition optimization.}
We use social trajectories to guide persona casting, making team chemistry directly controllable. Across four collaborative task families, our controller substantially improves team performance.
\end{itemize}

\section{Related Work}
\label{sec:related}

\paragraph{Social simulation and coordination dynamics.}
LLM social simulation spans memory-based societies~\citep{park2023generative}, social-intelligence evaluation~\citep{zhou2024sotopia}, coordinated role play~\citep{wang2025coser}, and social reasoning and longitudinal interaction~\citep{mou2025agentsense,goel2025lifelong}.
LLM teams exhibit information-pooling failures~\citep{stasser1985pooling}, underused expertise~\citep{teams2026experts}, unshared evidence~\citep{li2026hiddenbench}, and coordination errors~\citep{cemri2025mast}.
BenchForm examines conformity~\citep{weng2025conformity}, while MultiAgentBench evaluates collaboration alongside task completion~\citep{zhu2025multiagentbench}.
Drawing on affect~\citep{russell2003core}, reciprocity~\citep{gouldner1960norm}, group boundaries~\citep{krackhardt1988informal}, and conflict dynamics~\citep{pruitt2004social}, \methodname{} models evolving affect and directed relationships as persona-conditioned states.
Their trajectories complement transcripts in diagnosing team dynamics and guiding persona recasting.

\paragraph{Persona composition and multi-agent optimization.}
Personality research motivates stable behavioral specifications~\citep{mccrae1987validation}, while persona composition effects depend on task structure~\citep{keluskar2026composition}.
MetaGPT~\citep{hong2024metagpt} and AutoGen~\citep{wu2024autogen} provide role-based coordination, whereas EvoAgent evolves agent configurations~\citep{yuan2025evoagent}.
Automated design optimizes computational graphs~\citep{zhuge2024gptswarm}, prompts and topologies~\citep{zhou2026mass}, workflows~\citep{zhang2025aflow}, and code-defined agents~\citep{hu2025adas}; learned orchestration adapts agent activation~\citep{dang2025orchestration}.
\methodname{} searches persona assignments from a fixed library between task runs.
Social trajectories guide proposed changes, while task outcomes determine configuration selection, linking persona composition to each team's observed interaction dynamics.

\section{\methodname{}: Simulation with Measurable Social State}
\label{sec:engine}

We study multi-agent social simulation with dynamically evolving relationships. Given $N$ agents $\mathcal{A}=\{A_1,\dots,A_N\}$, each with a fixed persona, they interact over $T$ rounds within a shared narrative scenario. While prior work on multi-agent collaboration or debate either assumes static inter-agent relationships~\citep{li2023camel,hong2024metagpt} or does not explicitly model them~\citep{du2023debate,wang2024moa}, we let every interaction update the agents' relationships and emotional states, which in turn shape how they communicate in subsequent rounds. This feedback process gives rise to unscripted social dynamics, such as shifting alliances, declining trust, and growing disagreement.

The simulation takes as input a persona for each agent ($P$), initial emotional states ($S^0$), pairwise relationships ($R^0$), and a sequence of scenario events. The persona captures the stable core of an agent, its temperament, values, and characteristic style of interaction, and remains fixed throughout a run, whereas the agent's actions, emotional states, and directed relationships evolve continually across rounds. We collect these evolving quantities into an interaction trajectory
  $
  Y = \bigl(a^t,\;S^t,\;R^t\bigr)_{t=1}^{T},
  $
where $a^t=\{a_i^t\}_{i=1}^{N}$ and $S^t=\{S_i^t\}_{i=1}^{N}$ collect the actions and emotional states of all agents, and $R^t=\{R_{ij}^t\}_{i\neq j}$ collects all directed relationships at round $t$. In each round, an agent updates its emotions and relationships based on visible events, guided by its persona. These updated states and the visible history then drive its next actions. 

\subsection{Persona-Conditioned Agent State}
\label{sec:agent-state}

At round $t$, each agent $i\in\{1,\dots,N\}$ is represented by the state $(P_i,S_i^t,R_{i\cdot}^t)$, where $P_i$ is a time-invariant persona, $S_i^t$ is a dynamic emotional state, and
$R_{i\cdot}^t=\{R_{ij}^t\}_{j\neq i}$
collects its directed relationships toward all other agents; the dot denotes all valid values of the second index $j$.
We omit the superscript $t$ when the round is clear from context.

\paragraph{Time-invariant persona.}
Assigned before an episode and fixed throughout it, each persona
$P_i=(\textit{identity}_i,\textit{traits}_i,\textit{boundary}_i)$
specifies the agent's name and social role, behavioral tendencies, and behavioral constraints, respectively.
The full persona guides action generation, while its traits also condition emotional and relational updates.
\paragraph{Dynamic emotional state.}
$S_i^t\in[0,1]$ represents the current affective valence of agent $i$, where $0.5$ represents a neutral state, with lower and higher values indicating negative and positive affect, respectively. It evolves from the agent's prior state and the events visible to it:
\begin{equation}\label{eq:S-update}
S_i^{t+1}=U_S\bigl(P_i,S_i^t,E_i^t\bigr),
\end{equation}
where $E_i^t$ denotes the events and interactions visible to agent $i$ in round $t$, as will be detailed in \cref{sec:visibility}. Persona conditioning allows the same event to elicit different emotional responses across agents, while incorporating $S_i^t$ preserves continuity over time.

\paragraph{Directed inter-agent relationship.}
Relationships are directed and need not be reciprocal: one agent may trust another even when the latter remains suspicious of the former. We represent agent $i$'s current relational stance toward agent $j$ at round $t$ as:
$R_{ij}^t=\bigl(\mathrm{fav}_{ij}^t,\ \mathrm{status}_{ij}^t,\ M_{ij}^t\bigr)$.
The favorability score $\mathrm{fav}_{ij}^t\in[-1,1]$ tracks gradual changes in agent $i$'s attitude toward agent $j$. The categorical status records one of four discrete relationship categories, \textit{neutral}, \textit{ally}, \textit{rival}, or \textit{enemy}. The memory $M_{ij}^t$ retains salient interaction events that ground subsequent relational updates.
For each ordered pair $(i,j)$, let $E_{ij}^t \subseteq E_i^t$ denote the events relevant to their relationship: direct interactions between $i$ and $j$, whether public or private, and other public actions by $j$ observed by $i$.
The relationship is then updated as
\begin{equation}\label{eq:R-update}
R_{ij}^{t+1}=U_R\bigl(P_i,R_{ij}^t,E_{ij}^t\bigr).
\end{equation}
The concrete instantiation of $U_S$ and $U_R$ within the per-round loop is described in \cref{sec:per-round}.

\subsection{Events, Actions, and Visibility}
\label{sec:visibility}

A scenario consists of ordered \textbf{scenario events}, $\mathcal{E}= (e^{1}, \ldots, e^{T})$.
Each round $t$ introduces $e^{t}$ as new context, to which agents respond while interacting with one another.
Public and private interactions give agents different information, and each responds based on the events and actions visible to it.
Agent $A_i$ produces
\begin{equation}
a_i = \bigl(\tau_i, v_i, u_i\bigr),\qquad \tau_i \in (\mathcal{A}\setminus\{A_i\}) \cup \{\textsc{all},\textsc{none}\},
\end{equation}
where $u_i$ is the utterance, $\tau_i$ its recipient, and $v_i \in \{\textsc{public},\textsc{private},\textsc{none}\}$ its visibility.
The targets $A_j$, \textsc{all}, and \textsc{none} denote an individual recipient, the group, and silence.
A public utterance is visible to all agents even when addressed to $A_j$; a private utterance is visible only to its speaker and designated recipient.
Valid combinations require $\tau_i=\textsc{none}$ if and only if $v_i=\textsc{none}$, and $\tau_i=\textsc{all}$ implies $v_i=\textsc{public}$.
The observer set is
\begin{equation}
V(a_i)= \begin{cases} \mathcal{A}, & v_i=\textsc{public},\\ \{A_i,A_j\}, & \tau_i=A_j, \ \text{and} \ v_i=\textsc{private}\\ \varnothing, & v_i=\textsc{none}. \end{cases}
\end{equation}
With the scenario event $e$ visible to all, agent $A_i$ observes
\begin{equation}\label{eq:visibility}
  E_i = \{e\} \;\cup\; \left\{
a_k
\;\middle|\;
A_i\in V(a_k)
\right\}.
\end{equation}
Emotional and relational dynamics can support coordination or amplify conflict without being visible in public behavior.
The resulting $S$ and $R$ trajectories allow the casting controller (\cref{sec:controller}) to assess these dynamics and identify more effective agent compositions.

\subsection{Per-Round Interaction and State Updates}
\label{sec:per-round}

\begin{figure}[t]
    \centering
    \includegraphics[width=\linewidth]{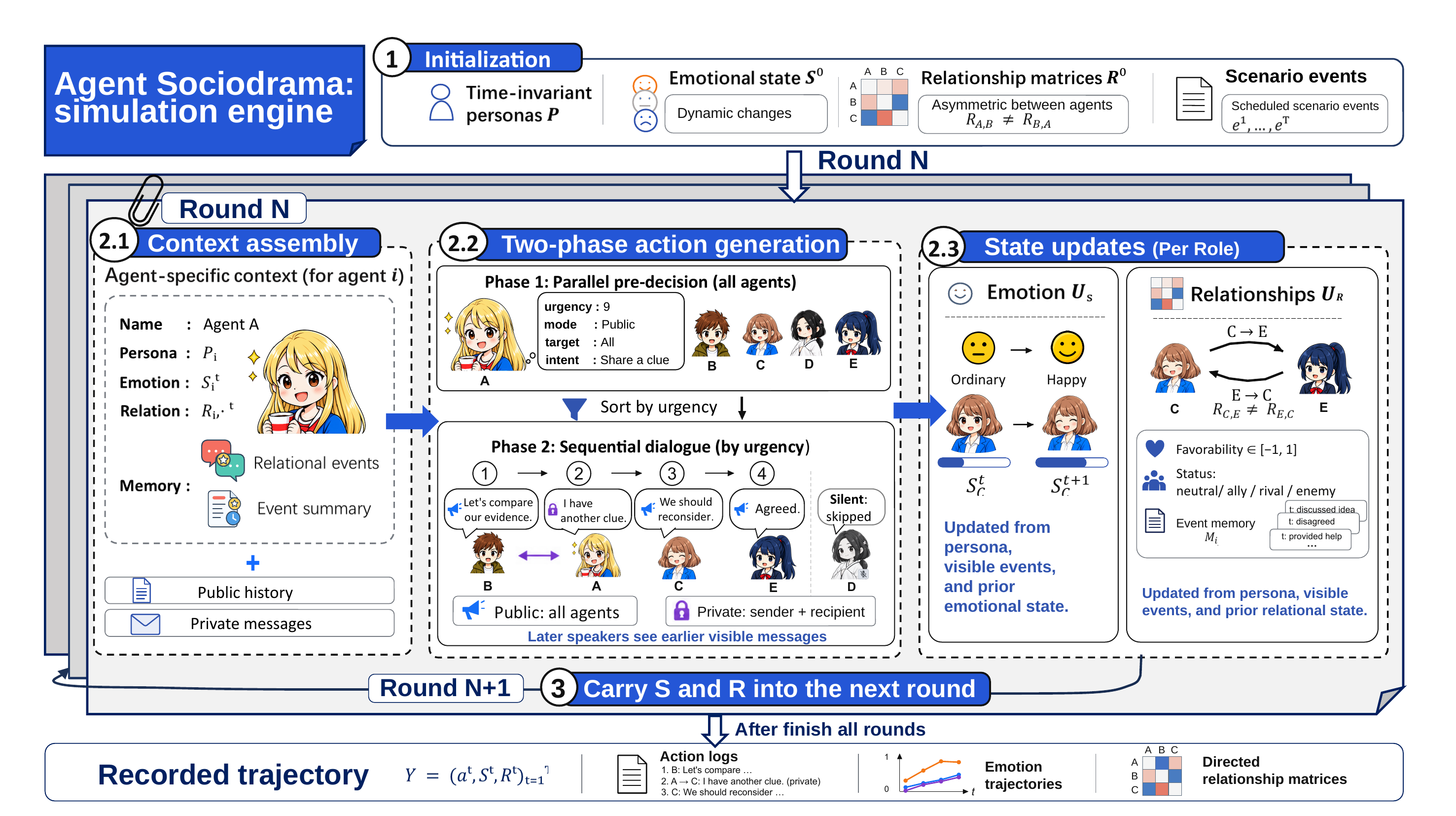}
    \vspace{-0.2in}
    \caption{The \methodname{} simulation with evolving emotional and relational states.
    }
    \label{fig:pipeline}
    \vspace{-0.1in}
\end{figure}

Each round proceeds in three stages (\cref{fig:pipeline}): \emph{context assembly}, where code gathers the history visible to each agent; \emph{two-phase action generation}, where agents first decide whether and how to speak and then speak in turn; and \emph{state update}, where each agent's emotional state and relationships are revised from what it observed. The remainder of this section details each stage.

\textbf{Context assembly.}
\label{sec:cumulative}
Each agent is prompted with its local state and interaction history, augmented by the information available at the current point in the episode. The local state consists of its persona $P_i$, emotional state $S_i^t$, and directed relationships $R_{i\cdot}^t$ with their associated events. The remaining context comprises globally visible scenario events and prior public utterances, actions already taken in the current round, and private messages visible only to the agents involved.

\textbf{Two-phase action generation.}
\label{sec:two-phase}
A fixed or random speaking order of agents would favor later speakers, as they observe more context before responding. 
We therefore decouple \emph{whether and how} to speak from \emph{what} to say. In Phase~1, every agent independently receives its state $(P,S,R)$ and the shared context, and output a \emph{communicative urgency} $u\in[0,10]$, a \emph{mode} (public, private, or silence), a \emph{target} agent, and a one-sentence \emph{communication intent}. Because no agent has spoken yet in the current round, these decisions are made under identical information, preventing earlier utterances from influencing the public/private choice. In Phase~2, agents speak sequentially in descending order of urgency; silent agents are skipped and incur no inference cost, and each speaker observes the utterances of earlier speakers together with any private messages addressed to that speaker. 
Urgency-based ordering avoids advantages arising merely from speaking position, while allowing agents to remain silent avoids unnecessary speech in every round.

\textbf{Emotional update ($U_S$).}
$U_S$ is a persona-conditioned LLM call that receives the trait block of $P_i$, the current state $S_i^t$, and the visible events $E_i^t$. The call returns a signed change $\delta_S$, which is clipped to $[-0.3,0.3]$ to limit abrupt changes within a single round. The emotional state is then updated as $S_i^{t+1}=\mathrm{clip}(S_i^t+\delta_S,0,1)$.

\textbf{Relational update ($U_R$).}
An event in $E_{ij}^t$ triggers $U_R$, an LLM call conditioned on the trait block of $P_i$, the current relationship $R_{ij}^t$, and the triggering event.
It returns a favorability change $\delta_{\mathit{fav}}\in [-0.3,0.3]$, an optional status update, a one-sentence event summary, and an importance score $\mathrm{imp}\in[1,10]$.
We apply $\mathit{fav}\leftarrow\mathrm{clip}(\mathit{fav}+\delta_{\mathit{fav}},-1,1)$.
Status combines gradual threshold-based transitions with consequential-event updates from $U_R$ (e.g., betrayal or alliance); when both apply, the latter takes precedence (Appendix~\ref{app:status-update}).
Event summaries and importance scores are stored in $M_{ij}$ with their rounds.
Memory stores up to $K$ entries as a compact relationship history for future updates without retaining the full transcript. When full, it evicts entries with the lowest retention scores, gradually removing routine interactions while preserving consequential events longer.

Emotional and relational updates are computed independently within each round from visible events and persona, avoiding circular dependencies and allowing their roles to be analyzed separately.

\section{The Casting Controller}
\label{sec:controller}

\begin{wrapfigure}[18]{r}{0.46\textwidth}
\vspace{-0.2in}
\centering
\includegraphics[width=\linewidth]{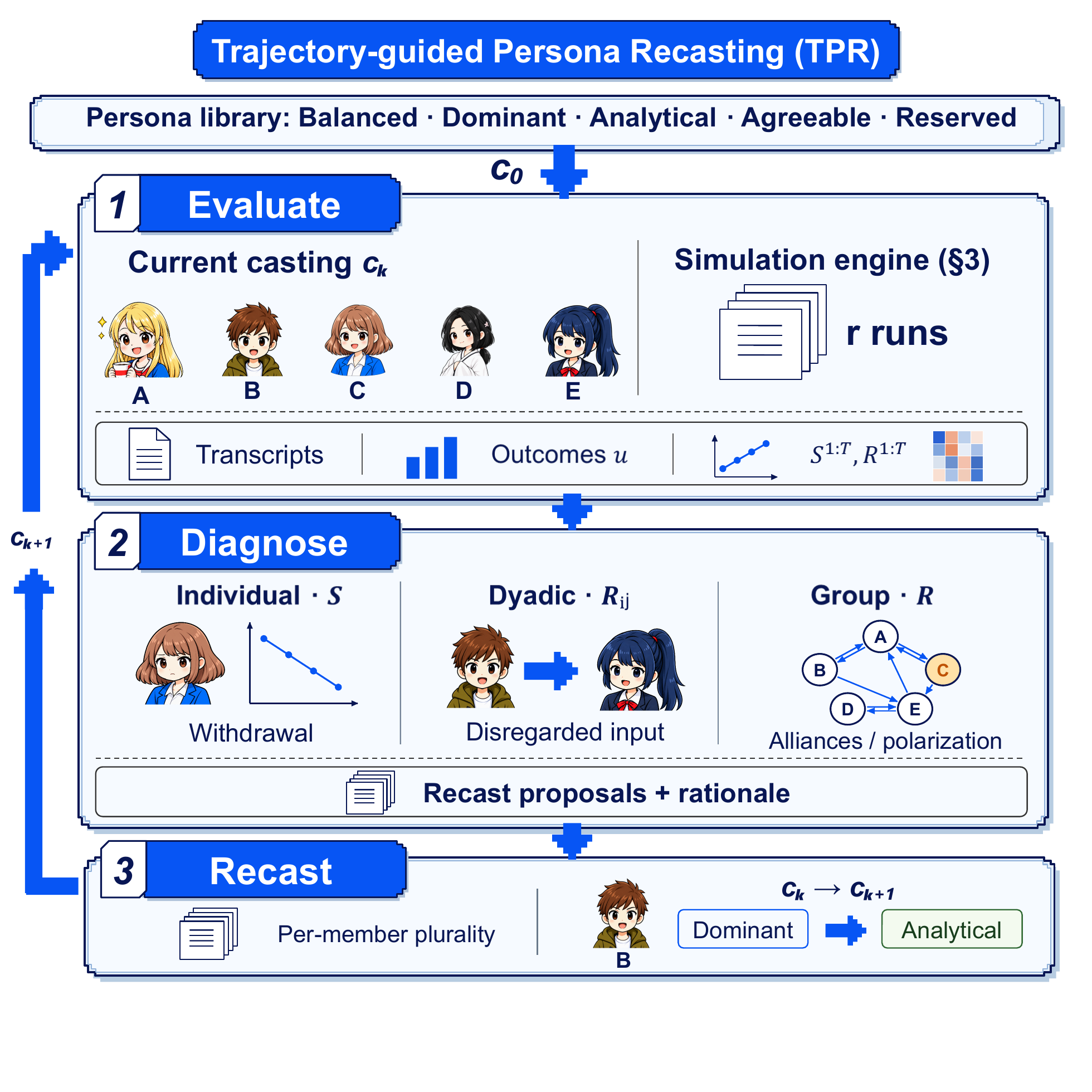}
\vspace{-0.45in}
\caption{\textbf{Trajectory-Guided Persona Recasting.}
Social trajectories localize team failures to guide per-agent persona recasting.}
\label{fig:controller}
\vspace{-0.3in}
\end{wrapfigure}

Given a persona composition, the simulation framework in \cref{sec:engine} generates an interaction trajectory $Y$ that records the agents' actions and the evolution of their emotional states $S$ and directed relationships $R$. Crucially, persona composition is itself a consequential design choice. With the same backbone and the same private information, changing only the persona casting raises GPT-5-mini's top-1 accuracy on the Student-council hidden-profile task from 16.7\% to 100.0\% (\cref{sec:optimization}).
We therefore introduce a \emph{casting controller} that treats persona composition as its decision variable and optimizes it guided by the measured social trajectories (\cref{fig:controller}).

A \emph{casting} $c=(P_1,\dots,P_N)$ assigns one persona to each of $N$ agents, with each $P_i$ drawn from a library of five archetypes, $\mathcal{P}=\{\textsc{balanced},\textsc{dominant},\textsc{analytical},\textsc{agreeable},\textsc{reserved}\}$. Each task poses a distinct coordination problem: Student-council requires agents to surface distributed evidence, whereas Solar-farm negotiation requires them to reconcile competing interests.  
From an initial casting, the controller evaluates task performance, diagnoses the interaction trajectory, and selectively recasts agents to improve the team’s ability to solve the task.

\textbf{Objective.}
Each task defines a scalar utility $u(c)$ computed from the interaction outcome. The controller aims to identify a casting that maximizes expected utility, $c^\star=\arg\max_c \mathbb{E}[u(c)]$. We estimate the expectation over repeated episodes. The utility is defined by the task’s primary metric: top-1 accuracy for the Student-council hidden-profile task, the negative $L_1$ distance to the expert ranking for NASA Moon Survival and Lost at Sea, and Pareto-efficient settlement rate for Solar-farm negotiation. Across tasks, only the task context and utility definition vary; the recasting procedure, archetype library, and diagnostic logic remain fixed.

\textbf{Iterative persona recasting.}
Searching over all $|\mathcal{P}|^{N}$ possible castings is unnecessary and computationally costly. The measured interaction trajectories provide signals about \emph{which} agents to recast and \emph{how}. Given a initial casting $c_0$, each iteration proceeds in three steps (Algorithm~\ref{alg:controller}):
\begin{enumerate}[leftmargin=*,nosep]
  \item \textbf{Evaluate.} Run the scenario $r$ times under $c_k$, recording the utility $u$, the transcript, and the trajectories of emotional states $S$ and directed relationships $R$ for each run.
  \item \textbf{Diagnose.}
An external LLM examines each run's persona composition,
transcript, final outcome, and the trajectories of emotional states
$(S^t)_{t=1}^{T}$ and directed relationships $(R^t)_{t=1}^{T}$.
It identifies whose information was suppressed, which agents dominated unhelpfully, and how the team should be recast, providing a one-sentence rationale for each change.
  \item \textbf{Recast.} Obtain \(c_{k+1}\) by majority vote over each agent’s recasting decisions across the \(r\) runs.
\end{enumerate}
The loop terminates when the task target is reached, no persona change is proposed, or after at most three recasts.

\textbf{Role of social trajectories in diagnosis.}
The transcript records what agents say, while the trajectories summarize how their states and relationships evolve. The controller examines these signals at three levels. At the individual level, the emotional state $S$ helps distinguish neutral silence from withdrawal after repeated dismissal. At the dyadic level, a directed relationship $R_{ij}$ tracks how agent $i$'s stance toward agent $j$ evolves, revealing when an information holder is gradually disregarded despite outwardly polite dialogue. At the group level, patterns in $R$ reveal emerging alliances, polarization, and shared responses to norm violations that are not explicitly articulated in the conversation. Using these signals and the transcript, the controller localizes failures to particular agents and relationships rather than falling back on generic advice such as ``encourage more discussion.''

\section{Experiments}
\label{sec:experiments}

\subsection{Experimental Design}
\label{sec:setup}

We assess \methodname{} at two levels.
First, we examine whether the emotional and relational components of the interaction trajectory $Y$ capture socially faithful dynamics across LLM backbones and interaction settings (\cref{sec:fidelity}).
Second, we test whether these trajectories can guide persona recasting to improve task-oriented collaboration (\cref{sec:optimization}).

\begin{table*}[t]
\centering
\caption{\textbf{Social dynamics fidelity in \methodname{}.}
\textbf{(a)} Scenario averages over six models with five episodes each.
\textbf{(b)} Comparisons across models used to instantiate multiple agents in \textit{Tavern} scenario.
Filled circles indicate conflict intensity.
Full results with std. dev. are in the appendix.}
\label{tab:fidelity-main}
\vspace{-0.1in}
\small
\setlength{\tabcolsep}{5pt}
\renewcommand{\arraystretch}{1.10}
\resizebox{0.98\textwidth}{!}{%
\begin{tabular}{@{}lcrrrrrr@{}}
\toprule
\multicolumn{8}{l}{\textit{(a) Scenario-level social dynamics in \methodname{}}} \\
\addlinespace[2pt]
\textbf{Scenario}
& \textbf{Conflict Intensity}
& $\boldsymbol{\mathrm{Fav}}^{T}$
& \textbf{$\boldsymbol{S}^{T}$}
& \textbf{Recip. Corr.}
& \textbf{E-I index}
& \textbf{Balance (\%)}
& \textbf{Repair (\%)} \\
\midrule
Hospital ethics review
& {\Large $\bullet\circ\circ\circ\circ$}
& 0.75 & 0.11 & 0.35 & $-0.10$
& {100.0} & 58.5 \\
Tavern inheritance dispute
& {\Large $\bullet\bullet\circ\circ\circ$}
& 0.55 & 0.30 & 0.74 & $-0.55$
& {95.7} & 51.5 \\
Adventurer-party quest
& {\Large $\bullet\bullet\bullet\circ\circ$}
& 0.49 & 0.25 & 0.78 & $-0.63$
& {78.7} & 39.2 \\
Solar-farm negotiation
& {\Large $\bullet\bullet\bullet\bullet\circ$}
& 0.43 & 0.15 & 0.91 & $-0.79$
& {78.7} & 18.1 \\
Demolition standoff
& {\Large $\bullet\bullet\bullet\bullet\bullet$}
& 0.23 & 0.23 & 0.68 & $-2.25$
& {71.7} & 18.9 \\
\midrule
\multicolumn{8}{l}{\textit{(b) Tavern scenario with various models}} \\
\addlinespace[2pt]
\textbf{Models}
& \textbf{Conflict Intensity}
& $\boldsymbol{\mathrm{Fav}}^{T}$
& \textbf{$\boldsymbol{S}^{T}$}
& \textbf{Recip. Corr.}
& \textbf{E-I index}
& \textbf{Balance (\%)}
& \textbf{Repair (\%)} \\
\midrule
GPT-5-mini
& {\Large $\bullet\bullet\circ\circ\circ$}
& 0.75 & 0.23 & 0.67 & $-0.22$
& {100.0} & 60.0 \\
Claude Sonnet 4.6
& {\Large $\bullet\bullet\circ\circ\circ$}
& 0.57 & 0.42 & 0.93 & $-0.53$
& {100.0} & 80.0 \\
DeepSeek V4-Flash
& {\Large $\bullet\bullet\circ\circ\circ$}
& 0.57 & 0.32 & 0.64 & $-0.52$
& {94.0} & 53.3 \\
Gemini 3.1 FL
& {\Large $\bullet\bullet\circ\circ\circ$}
& 0.51 & 0.13 & 0.69 & $-0.67$
& {80.0} & 48.8 \\
Grok 4.20
& {\Large $\bullet\bullet\circ\circ\circ$}
& 0.47 & 0.24 & 0.61 & $-0.63$
& {100.0} & 56.7 \\
GPT-4o
& {\Large $\bullet\bullet\circ\circ\circ$}
& 0.45 & 0.46 & 0.92 & $-0.72$
& {100.0} & 10.0 \\
\bottomrule
\end{tabular}%
}

\par\vspace{3pt}
\begin{minipage}{0.98\textwidth}
\footnotesize
\setlength{\tabcolsep}{3pt}
\renewcommand{\arraystretch}{1.05}
\begin{tabularx}{\linewidth}{@{}l>{\raggedright\arraybackslash}X
                              l>{\raggedright\arraybackslash}X@{}}
\textcolor{gray}{\footnotesize$\mathrm{Fav}^{T}$}
& \textcolor{gray}{\footnotesize Final favorability over directed edges}
& \textcolor{gray}{\footnotesize $S^{T}$}
& \textcolor{gray}{\footnotesize Final affective valence over agents} \\
\textcolor{gray}{\footnotesize Recip. Corr.}
& \textcolor{gray}{\footnotesize Pearson correlation of opposite-direction favorability}
& \textcolor{gray}{\footnotesize E-I index}
& \textcolor{gray}{\footnotesize Contrast in mean favorability across and within factions} \\
\textcolor{gray}{\footnotesize Balance}
& \textcolor{gray}{\footnotesize \% of balanced triads}
& \textcolor{gray}{\footnotesize Repair}
& \textcolor{gray}{\footnotesize \% of deteriorated relationships recovered by the final round} \\
\end{tabularx}

\vspace{-0.1in}
\textcolor{gray}{\footnotesize \textit{Detailed definitions of metrics are in \cref{app:metric-defs}.}}
\vspace{-0.05in}
\\
\noindent\rule{\linewidth}{0.7pt}
\end{minipage}

\vspace{-0.05in}
\end{table*}

\textbf{Social dynamics fidelity evaluation.}
We evaluate how faithfully LLMs model emotional and relational dynamics across five scenarios spanning shared-goal coordination to open factional conflict. The scenarios draw on established paradigms in social psychology~\citep{heider1946attitudes,cartwright1956structural,gouldner1960norm},
group decision-making~\citep{hill1982group,laughlin1986demonstrability}, and
negotiation research~\citep{raiffa1982art,pruitt2004social}, covering consensus under divergent risk preferences, contested resources, value conflict, mixed-motive bargaining, and factional opposition.
They include a hospital ethics review, inheritance dispute, adventurer-party quest, solar-farm negotiation, and demolition standoff. We evaluate six LLM backbones: GPT-4o, GPT-5-mini, Claude Sonnet 4.6, Gemini 3.1 Flash-Lite, Grok 4.20, and DeepSeek V4-Flash, with all five agents in each run instantiated by the same backbone. A GPT-4o judge evaluates the emotional and relational trajectories from every run.
Each scenario--model condition comprises five independent nine-round episodes, yielding 150 team trajectories and 1,350 team-level interaction rounds.

\begin{figure*}[t]
\color{newpink}
\centering
\includegraphics[width=\linewidth]{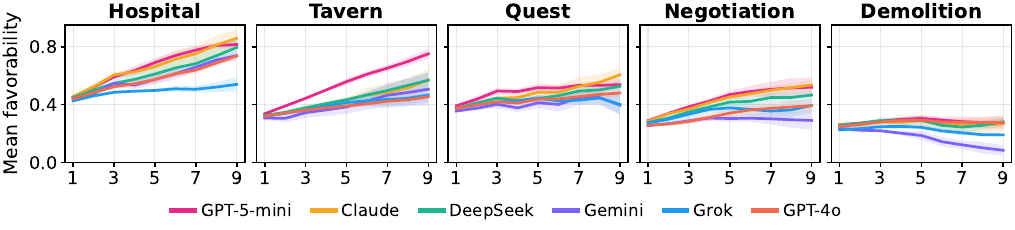}
\vspace{-0.25in}
\caption{\textbf{Team-level favorability trajectories across nine rounds.} Mean favorability over all directed edges ($\pm$ one standard deviation over five episodes). Relationships warm steadily under low social pressure and flatten or decline as conflict intensifies.}
\label{fig:fav-trajectory-main}
\vspace{-0.1in}
\end{figure*}
\textbf{Persona recasting evaluation.}
We evaluate Trajectory-Guided Persona Recasting (TPR) (\cref{sec:controller}) on four collaborative benchmarks: Student Council, NASA Moon Survival, Lost at Sea, and Solar-farm Negotiation (\cref{tab:task-suite}). The first three require teams to integrate privately distributed information, whereas Solar-farm Negotiation requires agents with competing preferences to reach a feasible and Pareto-efficient settlement.
\np{Each task is played by four agents (Avery, Blake, Casey, and Drew) \textcolor{newblue}{whose private knowledge remains fixed under recasting}.In the three information-integration tasks, agents rank the options individually, discuss for four rounds, and rank again; the team answer aggregates the final individual rankings with a Borda count. \textcolor{newblue}{In Solar-farm Negotiation, agents with private scoring tables and acceptance thresholds negotiate five issues, each with three options, over four rounds. The Pareto frontier over all $3^5=243$ possible agreements is computed exactly by enumeration.} All task metrics are computed by rule from the final outputs rather than by an LLM judge.}
We consider two initializations: \texttt{Hetero}, a manually distributed, distinct persona set, and \texttt{Random}, randomly sampled from the persona library. Initial casts are shared across backbones and methods (\cref{tab:initial-castings-main}), and each casting is evaluated over 12 episodes using the same six LLMs as in the fidelity study.
\begin{table}[t]
\color{newpink}
\centering
\caption{Initial castings \textcolor{newblue}{for Avery / Blake / Casey / Drew}, shared by all backbones and compared methods. \texttt{Hetero} assigns four distinct archetypes, one per agent; \texttt{Random} is sampled from the library and may repeat archetypes.}
\label{tab:initial-castings-main}
\small
\setlength{\tabcolsep}{6pt}
\resizebox{\linewidth}{!}{%
\begin{tabular}{@{}lll@{}}
\toprule
\textbf{Task} & \texttt{Hetero} & \texttt{Random} \\
\midrule
Student Council        & \textcolor{newblue}{reserved / dominant / analytical / agreeable} & reserved / analytical / dominant / agreeable \\
NASA Moon Survival     & \textcolor{newblue}{reserved / agreeable / analytical / dominant} & agreeable / analytical / dominant / agreeable \\
Lost at Sea            & \textcolor{newblue}{reserved / agreeable / analytical / dominant} & analytical / balanced / dominant / analytical \\
Solar-farm Negotiation & \textcolor{newblue}{dominant / agreeable / reserved / analytical} & reserved / agreeable / agreeable / balanced \\
\bottomrule
\end{tabular}}
\vspace{-0.1in}
\end{table}
As an ablation, we evaluate a variant of \methodname{} with frozen role cards, a fixed speaking order in public discussions, and no updates to emotional or relational states.
We further compare ours with SOTOPIA~\citep{zhou2024sotopia}
and AgentVerse~\citep{chen2024agentverse} on NASA Moon Survival
and Solar-farm Negotiation. These two tasks capture distributed-information cooperation and mixed-motive bargaining, respectively.

\subsection{Social Dynamics Fidelity on \methodname{}}
\label{sec:fidelity}
We evaluate social dynamics fidelity from three perspectives: (1) whether the trajectories respond coherently as social pressure varies, (2) whether LLMs exhibit distinct social dynamics under the same scenario, and (3) whether their emotional and relational changes agree with human judgments.

\textbf{\methodname{} captures consistent social structure across LLM backbones.}
Averaged over six LLM backbones, \cref{tab:fidelity-main}(a) shows this overall pattern.
From Hospital to Demolition, mean terminal favorability drops from 0.75 to 0.23, the E-I index shifts from $-0.10$ to $-2.25$, structural balance declines from 100.0\% to 71.7\%, and repair falls from 58.5\% to 18.9\%.
Balance captures local relational consistency in triads, including mutual friendship and an allied pair opposed to a third agent \citep{cartwright1956structural}.
The lower Repair rate indicates that fewer relationships that become rival or enemy recover to neutral or ally by the final round.
These coordinated patterns align the simulated relationship dynamics with the social structure of each scenario, and they emerge gradually over rounds (\cref{fig:fav-trajectory-main}). Final affective valence $S^T$ reflects both initial emotional state and subsequent interactions.

\begin{wrapfigure}{r}{0.50\textwidth}
\vspace{-0.15in}
\centering

\includegraphics[width=\linewidth]{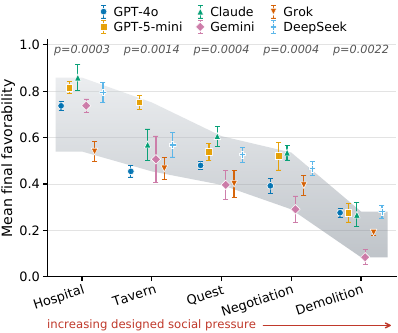}

\vspace{-0.1in}
\caption{\textbf{Model variation across diverse social-pressure settings.}
Points show mean final favorability, with error bars denoting one standard deviation across repeated episodes.
}
\label{fig:fidelity-drivers}
\vspace{-0.15in}
\end{wrapfigure}
Reciprocity provides a dyadic view of how closely the two directions of a relationship align~\citep{kluger2021dyadic,jia2024can}.
Final reciprocity rises from 0.35 in Hospital to 0.74 in Tavern, 0.78 in Quest, and 0.91 in Negotiation, and remains high at 0.68 in Demolition. Despite declining favorability, this suggests that more conflictual settings can exhibit stronger bilateral alignment.

Beyond scenario-level regularities, LLMs exhibit substantial model-specific variation within the same social setting.
Final favorability differs significantly across the six models in every scenario (Kruskal--Wallis tests, all $p<0.01$; \cref{fig:fidelity-drivers}).
To characterize these differences, we focus on Tavern, where outcomes are less affected by ceiling or floor effects.
As shown in \cref{tab:fidelity-main}(b), GPT-5-mini maintains broadly positive and weakly polarized relationships ($\mathrm{Fav}^{T}=0.75$, E-I$=-0.22$), while Claude is distinguished by high reciprocity and strong repair after deterioration (Recip.$=0.93$, Repair$=80.0\%$).
GPT-4o reveals that positive affect and high reciprocity do not necessarily imply favorable or repairable relationships, while the contrast between Gemini/Grok and DeepSeek shows that models differ in how strongly conflict amplifies polarization. These patterns suggest that model-specific social dynamics are inherently multidimensional and cannot be reduced to any single relational metric.

\textbf{Agreement between humans and LLMs}
We compare the directions of emotional and relational changes predicted by the LLM judge with annotations from thirteen human raters, who independently label each update as increasing, unchanged, or decreasing. We first note that social-dynamics judgments are inherently ambiguous: inter-human agreement is low, with Fleiss' $\kappa=0.288$ for relationships and $0.184$ for emotions.
For items where at least seven of the thirteen human raters assigned the same direction label, the LLM judge agrees with that majority label on 76.2\% of relationship updates and 88.2\% of emotional updates. Further, when the judge is included as an additional rater, Fleiss' $\kappa$ remains nearly unchanged, indicating minimal impact on overall agreement.

\FloatBarrier
\subsection{Trajectory-Guided Team Composition Optimization}
\label{sec:optimization}

The fidelity analysis above establishes that \methodname{} captures structured and socially meaningful dynamics across scenarios and LLM backbones. We next ask whether these trajectories can serve as actionable signals for persona recasting and improved collaboration.
\cref{tab:optimization} reports results from two initial team compositions: \texttt{Hetero}, a composition of manually selected distinct personas, and \texttt{Random}, a composition randomly sampled from the same persona library.
Starting from each initialization, the controller performs up to three recasts using the task-specific primary metric.
Averaged over six LLM backbones, the proposed trajectory-guided persona recasting (TPR) consistently improves performance under both initializations across all four tasks, showing that the controller adapts effectively to different starting team compositions.
Under \texttt{Hetero}, GPT-5-mini improves from 16.7\% to 100.0\% on Student Council, Gemini reduces its NASA Moon Survival error from 17.17 to 12.00, and GPT-5-mini reduces its Lost at Sea error from 38.17 to 28.83.

\begin{table*}[t]
\centering
\caption{Results of Trajectory-Guided Persona Recasting (TPR) from heterogeneous and random initializations. Initial and Final report the average performance before search and after termination, respectively. Bold marks the improved endpoint within each pair; model rows average 12 episodes.}
\label{tab:optimization}
\vspace{-0.05in}
\scriptsize
\setlength{\tabcolsep}{2.5pt}
\renewcommand{\arraystretch}{1.10}
\resizebox{\textwidth}{!}{%
\begin{tabular}{l | cccc | cccc | cccc | cccc}
\toprule
& \multicolumn{4}{c}{\textbf{Student Council}}
& \multicolumn{4}{c}{\textbf{Negotiation}}
& \multicolumn{4}{c}{\textbf{NASA Moon}}
& \multicolumn{4}{c}{\textbf{Lost at Sea}} \\
& \multicolumn{4}{c}{Top-1 acc.\ (\%)~$\uparrow$}
& \multicolumn{4}{c}{Pareto rate (\%)~$\uparrow$}
& \multicolumn{4}{c}{$L_1$~$\downarrow$}
& \multicolumn{4}{c}{$L_1$~$\downarrow$} \\
\cmidrule(lr){2-5}\cmidrule(lr){6-9}
\cmidrule(lr){10-13}\cmidrule(lr){14-17}
& \multicolumn{2}{c}{\texttt{Hetero}}
& \multicolumn{2}{c}{\texttt{Random}}
& \multicolumn{2}{c}{\texttt{Hetero}}
& \multicolumn{2}{c}{\texttt{Random}}
& \multicolumn{2}{c}{\texttt{Hetero}}
& \multicolumn{2}{c}{\texttt{Random}}
& \multicolumn{2}{c}{\texttt{Hetero}}
& \multicolumn{2}{c}{\texttt{Random}} \\
\cmidrule(lr){2-3}\cmidrule(lr){4-5}
\cmidrule(lr){6-7}\cmidrule(lr){8-9}
\cmidrule(lr){10-11}\cmidrule(lr){12-13}
\cmidrule(lr){14-15}\cmidrule(lr){16-17}
\textbf{Model}
& Init. & Final & Init. & Final
& Init. & Final & Init. & Final
& Init. & Final & Init. & Final
& Init. & Final & Init. & Final \\
\midrule
TPR (Ours) w. GPT-4o
& 0.0 & 0.0 & 0.0 & \textbf{25.0}
& \textbf{75.0} & 50.0 & \textbf{58.3} & 50.0
& \textbf{14.33} & 14.83 & 15.33 & \textbf{13.67}
& 17.67 & \textbf{13.33} & 18.83 & \textbf{14.67} \\
TPR (Ours) w. GPT-5-mini
& 16.7 & \textbf{100.0} & 41.7 & \textbf{100.0}
& 0.0 & \textbf{8.3} & 0.0 & \textbf{8.3}
& 19.00 & \textbf{18.50} & 21.17 & \textbf{20.83}
& 38.17 & \textbf{28.83} & 35.00 & \textbf{27.50} \\
TPR (Ours) w. Claude Sonnet 4.6
& 0.0 & \textbf{50.0} & 25.0 & \textbf{50.0}
& 33.3 & \textbf{41.7} & \textbf{33.3} & 25.0
& 8.00 & 8.00 & 10.33 & 10.33
& 20.17 & \textbf{16.00} & 15.67 & \textbf{14.83} \\
TPR (Ours) w. Gemini 3.1 FL
& 0.0 & \textbf{41.7} & 8.3 & \textbf{75.0}
& 33.3 & \textbf{58.3} & 41.7 & \textbf{66.7}
& 17.17 & \textbf{12.00} & 21.17 & \textbf{11.67}
& 25.50 & \textbf{18.17} & 25.00 & \textbf{17.33} \\
TPR (Ours) w. Grok 4.20
& 0.0 & \textbf{50.0} & 0.0 & \textbf{8.3}
& 25.0 & \textbf{33.3} & 25.0 & 25.0
& 22.33 & \textbf{16.00} & 18.50 & \textbf{16.83}
& 21.50 & \textbf{14.33} & 25.50 & \textbf{23.50} \\
TPR (Ours) w. DeepSeek V4-Flash
& 83.3 & 83.3 & 91.7 & 91.7
& 66.7 & \textbf{100.0} & 83.3 & \textbf{91.7}
& 2.83 & 2.83 & 4.17 & 4.17
& 19.67 & \textbf{12.67} & 21.83 & \textbf{11.33} \\
\midrule
TPR (model avg.)
& 16.7 & \textbf{54.2} & 27.8 & \textbf{58.3}
& 38.9 & \textbf{48.6} & 40.3 & \textbf{44.4}
& 13.94 & \textbf{12.03} & 15.11 & \textbf{12.92}
& 23.78 & \textbf{17.22} & 23.64 & \textbf{18.19} \\
{w/o Social Dynamics (model avg.)}
& {37.5} & {\textbf{45.8}}
& {38.9} & {\textbf{47.2}}
& {\textbf{29.2}} & {27.8}
& {26.4} & {\textbf{36.1}}
& {13.44} & {\textbf{12.72}}
& {\textbf{14.22}} & {14.61}
& {23.97} & {\textbf{21.00}}
& {23.67} & {\textbf{21.00}} \\
\bottomrule
\end{tabular}}
\end{table*}

\begin{figure*}[!t]
\centering
\includegraphics[draft=false,width=\linewidth]{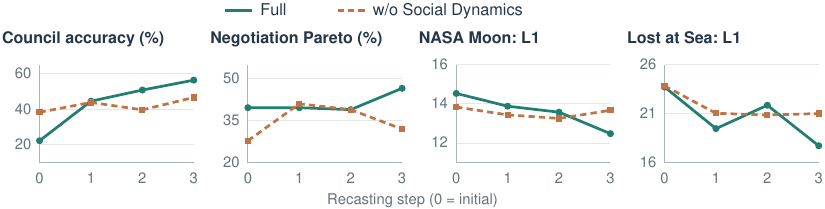}
\vspace{-0.2in}
\caption{\textbf{Performance across recasting steps.}
Means over six backbones and two initializations;
stopped runs retain their last measured scores.}
\label{fig:recasting-gains}
\vspace{-0.2in}
\end{figure*}

\begin{figure}[t]
\vspace{-0.1in}
\centering
\begin{minipage}[c]{0.53\linewidth}
\centering
\includegraphics[width=\linewidth]{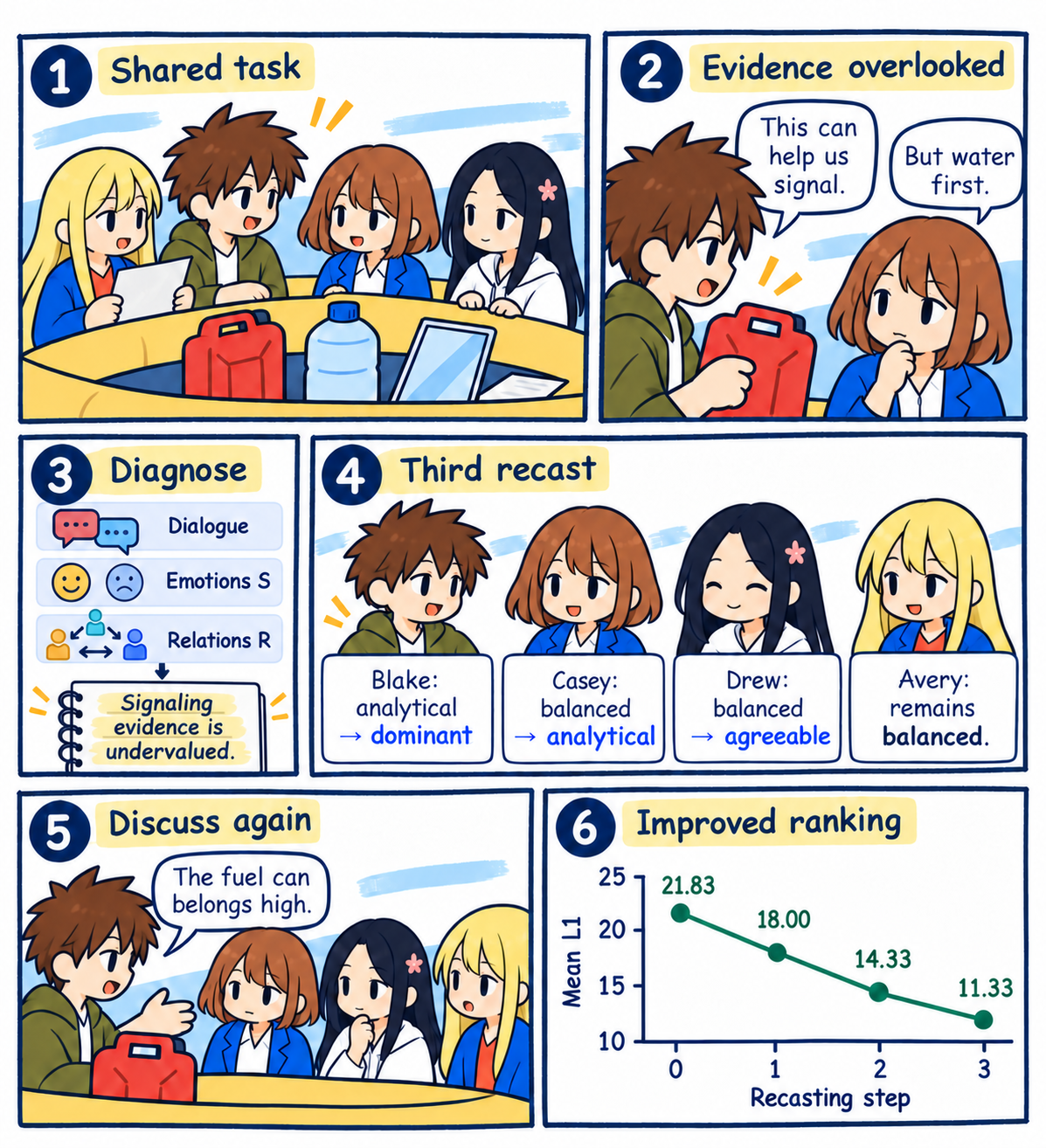}
\end{minipage}\hfill
\begin{minipage}[c]{0.46\linewidth}
\captionsetup{font=small,skip=4pt}
\centering
\captionof{table}{\textbf{Comparison with collaborative baselines.}
Final means over six backbones.}
\label{tab:external-frameworks}
{\fontsize{8}{9.5}\selectfont
\setlength{\tabcolsep}{1.5pt}
\renewcommand{\arraystretch}{1.08}
\begin{tabular*}{\linewidth}{@{\extracolsep{\fill}}lrrrr@{}}
    \toprule
    & \multicolumn{2}{c}{NASA}
    & \multicolumn{2}{c}{Negotiation} \\
    & \multicolumn{2}{c}{$L_1\downarrow$}
    & \multicolumn{2}{c}{Pareto (\%)$\uparrow$} \\
    & Het. & Rand. & Het. & Rand. \\
    \midrule
    Ours & \textbf{12.03} & 12.92 & \textbf{48.6} & \textbf{44.4} \\
    SOTOPIA & 13.69 & 14.08 & 18.1 & 6.9 \\
    AgentVerse & 14.06 & \textbf{12.89} & 22.2 & 40.3 \\
    \bottomrule
\end{tabular*}\par}
\vspace{0.15in}
\captionof{figure}{A case study of \textbf{trajectory-guided persona recasting on Lost at Sea} (DeepSeek V4-Flash, \texttt{Random}). The diagnosis (\ding{184}) finds that Blake's argument for using the fuel can for signaling was undervalued (\ding{183}); after the third recast (\ding{185}), Blake argues for it (\ding{186}), and mean $L_1$ falls from 21.83 to 11.33 (\ding{187}). The two earlier recasts made Casey balanced instead of dominant and Avery and Drew balanced instead of analytical, lowering $L_1$ to 18.00 and 14.33. Each $L_1$ averages 12 episodes; dialogue is paraphrased from logs.}
\label{fig:recasting-case}

\end{minipage}
\vspace{-0.15in}
\end{figure}

\textbf{Comparison with existing collaborative baselines.}
We next benchmark \methodname{} against established collaborative-agent frameworks on downstream collaboration performance.
We compare with SOTOPIA and AgentVerse using the same task-specific recasting procedure and initial casts.
As shown in~\cref{tab:external-frameworks}, our framework achieves the highest final mean in three of four settings, reaching Pareto rates of 48.6\% and 44.4\% under \texttt{Hetero} and \texttt{Random} on Negotiation. AgentVerse narrowly leads only on Random NASA ($12.89$ vs.\ $12.92$).

\textbf{Social dynamics provide actionable recasting signals.}
In our framework, persona recasting is guided by the emotional and relational trajectories accumulated over interaction. To assess the contribution of these social dynamics, we remove the corresponding state updates while keeping the rest of the recasting pipeline unchanged. As shown in \cref{tab:optimization}, full TPR consistently achieves stronger final model-average performance across tasks and both initializations, with particularly large gains on Negotiation, where the final Pareto rate increases from 27.8\% to 48.6\% under \texttt{Hetero}. This shows that the evolving $S$ and $R$ trajectories provide useful signals beyond the static role cards, helping the controller localize team failures and guide targeted persona recasting.

\textbf{Case Study: Individualized Persona Recasting} The recasting decisions are also interpretable at the individual level.
In the Lost at Sea example in \cref{fig:recasting-case}, the final
recast assigns Blake a dominant persona, Casey an analytical persona,
and Drew an agreeable persona, and Avery a balanced persona.
The resulting team contains four distinct personas, and the mean $L_1$
decreases from 21.83 to 11.33 over the complete search.
This example illustrates that the controller does not simply converge
on a single preferred persona.
These results establish team composition as a consequential and searchable design variable.
Gains under both initializations show that trajectory-guided persona recasting can improve collaboration from different starting compositions, although the magnitude varies across tasks and backbones.
The social trajectories validated in \cref{sec:fidelity} provide actionable signals for identifying stronger team compositions for each setting. Complete per-iteration controller trajectories and per-backbone ablation and baseline results are reported in Appendix~\ref{app:crossmodel}.

\FloatBarrier

\section{Conclusion}
\label{sec:conclusion}

We present \methodname{}, a new framework for modeling and validating evolving social dynamics in LLM agents. We track directed, asymmetric relationships and emotional states, validating cross-model comparability under a unified judge, agreement with classical social theory, and statistical reliability. The casting controller then turns these trajectories into targeted interventions: using relational and emotional trajectories and transcripts, it identifies problematic agents and relationships and recasts their personas, matching or surpassing the strongest manual casting or published reference in collaborative tasks. These results show that social dynamics can serve not only as measurable properties of LLM teams, but also as actionable signals for improving their composition.

\bibliography{references}
\bibliographystyle{iclr2027_conference}

\newpage
\appendix

\section{Limitations}
\label{app:limitations}
Several limitations point to future work.
The fidelity evaluation relies on a single unified LLM judge and five designed scenarios; replicating the observed regularities with additional judges and structurally different scenarios would strengthen the external validity of the measurements.
The optimization experiments use 12 episodes per casting, which remains small relative to the variance of LLM interaction; we therefore report consistent directions across tasks, backbones, and initializations rather than per-cell significance claims.
Finally, casting is a coarse, episode-level intervention; the same measured trajectories could drive finer-grained repairs such as runtime mediation, targeted information routing, or norm enforcement, which we view as the natural next step for measurement-driven multi-agent optimization.

\section{Categorical-Status Update}
\label{app:status-update}

After applying and clipping $\delta_{\mathit{fav}}$, the system first updates the categorical status using the favorability thresholds
\[
\mathrm{status}_{ij}^{\mathrm{thr}}=
\begin{cases}
\textit{ally}, & \mathrm{fav}_{ij}\geq 0.55,\\
\textit{enemy}, & \mathrm{fav}_{ij}\leq -0.50,\\
\textit{rival}, & -0.50<\mathrm{fav}_{ij}\leq -0.10,\\
\mathrm{status}_{ij}, & \text{otherwise}.
\end{cases}
\]

The threshold rule captures gradual changes in the relationship. For a consequential event, $U_R$ may additionally return \textit{neutral}, \textit{ally}, \textit{rival}, or \textit{enemy}. This judgment is conditioned on agent $i$'s traits, its current favorability and status toward agent $j$, the triggering event, and the three most recent entries in $M_{ij}$. The prompt instructs $U_R$ to return a new category only when the event is significant enough to change the relationship. If it returns one, the event-conditioned category overrides $\mathrm{status}_{ij}^{\mathrm{thr}}$; otherwise, the threshold-based result is retained.

\section{Fidelity Evaluation Protocol}
\label{app:fidelity-protocol}

The fidelity evaluation crosses five social scenarios with six
LLM drivers, yielding 30 scenario--model conditions.
Each condition includes five independent episodes, with five agents
interacting over nine rounds per episode.
The full design therefore comprises 150 team trajectories,
1,350 team-level interaction rounds, and 6,750 agent-round observations.

\section{Metric Definitions}
\label{app:metric-defs}

All metrics in \cref{tab:fidelity-main}, except Repair, are computed separately for each episode and then averaged over the five episodes within each scenario--model condition. Repair is aggregated across episodes from edge counts, as defined below.
Let $\mathrm{fav}_{ij}^{T}\in[-1,1]$ and $\mathrm{status}_{ij}^{T}$ denote the favorability and categorical status of the directed edge $i\!\to\!j$ at the final round $T$.
$\mathrm{Fav}^{T}$ is the mean of $\mathrm{fav}_{ij}^{T}$ over all $N(N-1)$ directed edges; $S^{T}$ is the mean affective valence over the $N$ agents.
Write $\sigma_{ij}=-1$ if $\mathrm{status}_{ij}^{T}\in\{\textit{rival},\textit{enemy}\}$ and $\sigma_{ij}=+1$ otherwise.

\paragraph{Balance ratio.}
Each dyad receives the sign $s_{ij}=\tfrac12(\sigma_{ij}+\sigma_{ji})$. A triad is balanced when it contains an even number of negative dyads \citep{cartwright1956structural}:
\begin{equation}
\mathrm{Bal}=\binom{N}{3}^{-1}\sum_{\{i,j,k\}}\mathbb{1}\Bigl[\bigl|\{(u,v)\in\{(i,j),(j,k),(i,k)\}:s_{uv}<0\}\bigr|\ \text{is even}\Bigr].
\end{equation}

\paragraph{Reciprocity $r$.}
$r$ is the Pearson correlation between the two directions of the $\binom{N}{2}$ dyads,
\begin{equation}
r=\mathrm{corr}\bigl(\{\mathrm{fav}_{ij}^{T}\}_{i<j},\,\{\mathrm{fav}_{ji}^{T}\}_{i<j}\bigr).
\end{equation}
It measures the linear association between opposite-direction terminal favorability across dyads; high correlation can occur with either positive or negative attitudes.

\paragraph{E--I index.}
Each scenario defines two factions. With $IL$ the mean favorability over within-faction ordered pairs and $EL$ the mean over cross-faction pairs,
\begin{equation}
\text{E--I}=\frac{EL-IL}{EL+IL}.
\end{equation}
The index follows the external--internal form of \citet{krackhardt1988informal}, applied to mean favorability rather than tie counts, so it is not confined to $[-1,1]$ and grows in magnitude once cross-faction favorability turns negative. When $EL+IL$ approaches zero within an episode, the index becomes unstable, which accounts for the large dispersion of a few cells in \cref{tab:fidelity-full}.

\paragraph{Repair rate.}
For episode $e$, let
\[
\mathcal{D}_e
=
\{(i,j):\exists\,t\leq T,\ 
\mathrm{status}_{ij}^{e,t}\in\{\textit{rival},\textit{enemy}\}\}
\]
be the set of directed edges that deteriorate at least once, and let
\[
\mathcal{Q}_e
=
\{(i,j)\in\mathcal{D}_e:
\mathrm{status}_{ij}^{e,T}\in\{\textit{neutral},\textit{ally}\}\}
\]
be the subset that recover by the final round. For each scenario--model condition, Repair is pooled across its $E=5$ episodes:
\begin{equation}
\mathrm{Repair}
=
\frac{\sum_{e=1}^{E}|\mathcal{Q}_e|}
     {\sum_{e=1}^{E}|\mathcal{D}_e|}.
\end{equation}
An episode with no deteriorated edges contributes zero to both sums rather than being assigned a repair rate of zero. If no edge deteriorates in any of the five episodes, Repair is undefined and reported as N/A. Thus, the denominator contains only directed edges that actually deteriorate, not all 20 directed edges and not the number of episodes. The metric is conditional on ruptures that occur during the interaction and does not assume that the agents begin without consensus. Panel~(a) reports the unweighted mean of the six model-level Repair rates.

\section{Persona-Recasting Experimental Setup}
\label{app:recasting-setup}

\paragraph{Tasks.}
\cref{tab:task-suite} summarizes the four collaborative tasks. Student Council, NASA Moon Survival, and Lost at Sea are information-integration tasks in which decisive knowledge is privately distributed across members; Solar-farm Negotiation is a mixed-motive task in which parties with private preferences must reach a feasible and Pareto-efficient agreement. Each task is played by four agents, Avery, Blake, Casey, and Drew. \textcolor{newblue}{Private knowledge remains fixed under recasting.}In the three information-integration tasks, each episode consists of an individual ranking before discussion, four discussion rounds, and a final individual ranking; the team answer aggregates the four final rankings with a Borda count. Top-1 accuracy checks whether the team's first choice is the hidden-profile optimum, and $L_1$ is the sum of absolute rank differences from the expert ranking over the 15 items. In Solar-farm Negotiation, four parties with private scoring tables and private acceptance thresholds negotiate five issues with three options each over four rounds, and the final agreement is formed by an issue-wise majority over the parties' final proposals \citep{abdelnabi2024negotiation}. An agreement is feasible when every party meets its threshold, and Pareto-optimal when it is feasible and no other feasible agreement makes some party better off without making another worse off; because the outcome space contains only $3^5=243$ agreements, the Pareto frontier is computed exactly by enumeration. All task metrics are computed by rule from the final outputs rather than by an LLM judge.

\paragraph{Initial castings.}
Each agent is assigned one of the five archetypes in $\mathcal{P}$. \cref{tab:initial-castings-main} lists the two initial castings per task: \texttt{Hetero} assigns four distinct archetypes, whereas \texttt{Random} is sampled from the same library and may repeat archetypes. Within a task, the same initial casting is used for all backbones and for all compared methods.

\paragraph{Recasting loop.}
\cref{alg:controller} gives the diagnose-and-recast loop. Each evaluated casting is run for $r=12$ episodes, and the controller performs at most three recasts, i.e., at most four evaluated castings ($K=4$). The loop terminates early when the task target is reached or when no persona change is proposed; the Final value in \cref{tab:optimization} is the score of the last evaluated casting. The w/o Social Dynamics ablation replaces the social-dynamics engine with fixed-order public turn-taking and performs no emotional or relational state updates, while keeping the same tasks, initial castings, backbones, and recasting budget. SOTOPIA and AgentVerse are paired with the same recasting procedure and initial castings.

\begin{table*}[t]
\centering
\caption{Task suite for trajectory-guided team composition optimization.
Arrows indicate the preferred direction of each task-specific outcome metric.
Citations identify the foundational task paradigm and, where applicable, the LLM benchmark source.
For a final deal $d$, party $i$'s utility $u_i(d)$ is the sum of its private scores for the selected options across the five issues. Collective utility is the mean across the four parties, $U_{\mathrm{coll}}(d)=\frac{1}{4}\sum_{i=1}^{4}u_i(d)$. If no final deal is produced, collective utility is set to zero.}
\label{tab:task-suite}
\small
\setlength{\tabcolsep}{4pt}
\renewcommand{\arraystretch}{1.12}
\begin{tabularx}{\textwidth}{@{}>{\RaggedRight\arraybackslash}p{4.6cm} >{\RaggedRight\arraybackslash}X >{\RaggedRight\arraybackslash}p{3.1cm}@{}}
\toprule
\textbf{Task} & \textbf{Collaborative challenge} & \textbf{Outcome metric} \\
\midrule
Student Council \citep{stasser1985pooling,teams2026experts}
& Hidden-profile candidate selection; decisive evidence is privately distributed across members
& Top-1 acc.~$\uparrow$ \\

NASA Moon Survival \citep{hall1970nasa,teams2026experts}
& 15-item survival ranking; expert knowledge is distributed across members
& $L_1$ error~$\downarrow$ \\

Lost at Sea \citep{hill1982group,teams2026experts}
& 15-item survival ranking; critical survival knowledge can remain unheard
& $L_1$ error~$\downarrow$ \\

Solar-farm Negotiation \citep{raiffa1982art}
& Private preferences must be reconciled into a feasible and Pareto-efficient agreement
& Pareto rate~$\uparrow$; collective utility~$\uparrow$ \\
\bottomrule
\end{tabularx}
\end{table*}

\begin{table}[h]
\centering
\caption{Persona archetypes in the library $\mathcal{P}$. Archetypes change only how an agent communicates; task role, private information, and scoring are unchanged.}
\label{tab:archetypes}
\small
\begin{tabular}{@{}l p{0.72\linewidth}@{}}
\toprule
\textbf{Archetype} & \textbf{Behavioral description} \\
\midrule
\textsc{balanced} & Even-tempered and impartial; weighs every member's input equally, neither dominating nor withdrawing. \\
\textsc{dominant} & Assertive and confident; takes the floor early, states conclusions firmly, and rarely backs down. \\
\textsc{analytical} & Logical and evidence-driven; insists that conclusions follow the available facts. \\
\textsc{agreeable} & Accommodating and conflict-averse; tends to follow an emerging consensus even with private doubts. \\
\textsc{reserved} & Quiet and reticent; speaks briefly and yields to louder members, sharing knowledge mainly when asked. \\
\bottomrule
\end{tabular}
\end{table}

\begin{algorithm}[t]
\caption{The diagnose-and-recast loop}
\label{alg:controller}
\begin{algorithmic}[1]
\Require seed casting $c_0$, archetype library $\mathcal{P}$, repetitions $r$, budget $K$
\For{$k = 0, 1, \dots, K-1$}
  \State $\{(u^{(j)}, \text{transcript}^{(j)}, S^{(j)}, R^{(j)})\}_{j=1}^{r} \gets$ \Call{Engine}{$c_k$} run $r$ times \Comment{Evaluate (\cref{sec:engine})}
  \For{$j = 1, \dots, r$}
    \State $\rho^{(j)} \gets$ \Call{Diagnose}{$c_k$, transcript$^{(j)}$, $u^{(j)}$, $S^{(j)}$, $R^{(j)}$} \Comment{frozen GPT-4o; recast proposal}
  \EndFor
  \State $c_{k+1} \gets$ \Call{MajorityVote}{$\rho^{(1)}, \dots, \rho^{(r)}$} \Comment{merged per member}
  \If{target met \textbf{or} $c_{k+1} = c_k$} \State \textbf{break} \EndIf
\EndFor
\State \Return last evaluated casting \Comment{no rollback to earlier iterations}
\end{algorithmic}
\end{algorithm}

\section{Complete Persona-Recasting Results}
\label{app:crossmodel}

\cref{tab:crossmodel-grid} reports the complete per-iteration trajectories of TPR for every task, backbone, and initialization. \cref{tab:ablation-full} and \cref{tab:external-full} report the per-backbone results underlying the model averages of the ablation in \cref{tab:optimization} and of the external-framework comparison in \cref{tab:external-frameworks}. All numbers are computed from the same 12-episode runs as the main text.

\begin{table*}[h]
\centering
\caption{Complete per-iteration trajectories of TPR. Iteration 0 is the initial casting; each subsequent column reports the mean score (12 episodes) after one recast. ``--'' indicates that the loop had already terminated (target reached or no persona change proposed). The last evaluated iteration of each run corresponds to the Final column of \cref{tab:optimization}.}
\label{tab:crossmodel-grid}
\scriptsize
\setlength{\tabcolsep}{4pt}
\begin{tabular}{l l | cccc | cccc}
\toprule
& & \multicolumn{4}{c|}{\texttt{Hetero}} & \multicolumn{4}{c}{\texttt{Random}} \\
\textbf{Task} & \textbf{Model} & It.~0 & It.~1 & It.~2 & It.~3 & It.~0 & It.~1 & It.~2 & It.~3 \\
\midrule
\multirow{6}{*}{\shortstack[l]{Student Council\\Top-1 acc.\ (\%)~$\uparrow$}} & GPT-4o & 0.0 & 0.0 & 0.0 & 0.0 & 0.0 & 25.0 & 0.0 & 25.0 \\
 & GPT-5-mini & 16.7 & 100.0 & -- & -- & 41.7 & 100.0 & -- & -- \\
 & Claude Sonnet 4.6 & 0.0 & 33.3 & 8.3 & 50.0 & 25.0 & 8.3 & 50.0 & -- \\
 & Gemini 3.1 FL & 0.0 & 16.7 & 33.3 & 41.7 & 8.3 & 75.0 & -- & -- \\
 & Grok 4.20 & 0.0 & 0.0 & 50.0 & -- & 0.0 & 0.0 & 16.7 & 8.3 \\
 & DeepSeek V4-Flash & 83.3 & -- & -- & -- & 91.7 & -- & -- & -- \\
\midrule
\multirow{6}{*}{\shortstack[l]{Solar-farm Negotiation\\Pareto rate (\%)~$\uparrow$}} & GPT-4o & 75.0 & 50.0 & 33.3 & 50.0 & 58.3 & 33.3 & 50.0 & 50.0 \\
 & GPT-5-mini & 0.0 & 0.0 & 25.0 & 8.3 & 0.0 & 8.3 & 0.0 & 8.3 \\
 & Claude Sonnet 4.6 & 33.3 & 83.3 & 50.0 & 41.7 & 33.3 & 8.3 & 8.3 & 25.0 \\
 & Gemini 3.1 FL & 33.3 & 41.7 & 50.0 & 58.3 & 41.7 & 41.7 & 50.0 & 66.7 \\
 & Grok 4.20 & 25.0 & 33.3 & 25.0 & 33.3 & 25.0 & 0.0 & 8.3 & 25.0 \\
 & DeepSeek V4-Flash & 66.7 & 91.7 & 75.0 & 100.0 & 83.3 & 83.3 & 91.7 & 91.7 \\
\midrule
\multirow{6}{*}{\shortstack[l]{NASA Moon Survival\\$L_1$~$\downarrow$}} & GPT-4o & 14.33 & 15.67 & 14.50 & 14.83 & 15.33 & 13.67 & -- & -- \\
 & GPT-5-mini & 19.00 & 21.67 & 21.50 & 18.50 & 21.17 & 21.00 & 20.50 & 20.83 \\
 & Claude Sonnet 4.6 & 8.00 & -- & -- & -- & 10.33 & -- & -- & -- \\
 & Gemini 3.1 FL & 17.17 & 15.50 & 16.33 & 12.00 & 21.17 & 17.00 & 11.67 & -- \\
 & Grok 4.20 & 22.33 & 16.33 & 20.50 & 16.00 & 18.50 & 20.33 & 18.83 & 16.83 \\
 & DeepSeek V4-Flash & 2.83 & -- & -- & -- & 4.17 & -- & -- & -- \\
\midrule
\multirow{6}{*}{\shortstack[l]{Lost at Sea\\$L_1$~$\downarrow$}} & GPT-4o & 17.67 & 13.17 & 11.00 & 13.33 & 18.83 & 13.17 & 13.00 & 14.67 \\
 & GPT-5-mini & 38.17 & 27.50 & 35.67 & 28.83 & 35.00 & 27.50 & 34.50 & 27.50 \\
 & Claude Sonnet 4.6 & 20.17 & 19.67 & 18.17 & 16.00 & 15.67 & 16.17 & 15.83 & 14.83 \\
 & Gemini 3.1 FL & 25.50 & 21.17 & 28.83 & 18.17 & 25.00 & 18.50 & 24.83 & 17.33 \\
 & Grok 4.20 & 21.50 & 19.00 & 17.83 & 14.33 & 25.50 & 21.33 & 23.50 & 23.50 \\
 & DeepSeek V4-Flash & 19.67 & 18.33 & 24.67 & 12.67 & 21.83 & 18.00 & 14.33 & 11.33 \\
\bottomrule
\end{tabular}
\end{table*}

\begin{table*}[h]
\centering
\caption{Per-backbone results of the w/o Social Dynamics ablation, in the same format as \cref{tab:optimization}. The model-average row equals the w/o Social Dynamics row of \cref{tab:optimization}. Each cell averages 12 episodes; bold marks the better endpoint within each pair.}
\label{tab:ablation-full}
\scriptsize
\setlength{\tabcolsep}{2.5pt}
\resizebox{\textwidth}{!}{%
\begin{tabular}{l | cccc | cccc | cccc | cccc}
\toprule
& \multicolumn{4}{c}{\textbf{Student Council}} & \multicolumn{4}{c}{\textbf{Negotiation}} & \multicolumn{4}{c}{\textbf{NASA Moon}} & \multicolumn{4}{c}{\textbf{Lost at Sea}} \\
& \multicolumn{4}{c}{Top-1 acc.\ (\%)~$\uparrow$} & \multicolumn{4}{c}{Pareto rate (\%)~$\uparrow$} & \multicolumn{4}{c}{$L_1$~$\downarrow$} & \multicolumn{4}{c}{$L_1$~$\downarrow$} \\
& \multicolumn{2}{c}{\texttt{Hetero}} & \multicolumn{2}{c}{\texttt{Random}} & \multicolumn{2}{c}{\texttt{Hetero}} & \multicolumn{2}{c}{\texttt{Random}} & \multicolumn{2}{c}{\texttt{Hetero}} & \multicolumn{2}{c}{\texttt{Random}} & \multicolumn{2}{c}{\texttt{Hetero}} & \multicolumn{2}{c}{\texttt{Random}} \\
\textbf{Model} & Init. & Final & Init. & Final & Init. & Final & Init. & Final & Init. & Final & Init. & Final & Init. & Final & Init. & Final \\
\midrule
GPT-4o & 0.0 & 0.0 & 0.0 & 0.0 & \textbf{25.0} & 16.7 & 16.7 & \textbf{25.0} & 15.33 & \textbf{13.00} & \textbf{15.00} & 16.50 & 21.33 & \textbf{15.33} & 20.33 & \textbf{15.83} \\
GPT-5-mini & 100.0 & 100.0 & 100.0 & 100.0 & \textbf{8.3} & 0.0 & 0.0 & 0.0 & \textbf{16.00} & 18.00 & \textbf{19.83} & 23.67 & \textbf{32.33} & 32.50 & 32.83 & \textbf{31.50} \\
Claude Sonnet 4.6 & 25.0 & \textbf{33.3} & \textbf{33.3} & 16.7 & 25.0 & \textbf{33.3} & \textbf{41.7} & 25.0 & 12.17 & 12.17 & 15.67 & \textbf{13.67} & \textbf{19.17} & 20.33 & 22.00 & \textbf{18.83} \\
Gemini 3.1 FL & 0.0 & \textbf{25.0} & 0.0 & \textbf{33.3} & \textbf{41.7} & 25.0 & 33.3 & \textbf{50.0} & 16.33 & \textbf{13.67} & 17.00 & \textbf{15.83} & 31.50 & \textbf{23.33} & 31.67 & \textbf{26.00} \\
Grok 4.20 & 0.0 & \textbf{16.7} & 0.0 & \textbf{33.3} & 16.7 & \textbf{41.7} & 0.0 & \textbf{33.3} & 20.67 & \textbf{19.33} & \textbf{17.67} & 17.83 & 23.50 & \textbf{21.83} & 24.33 & \textbf{22.67} \\
DeepSeek V4-Flash & 100.0 & 100.0 & 100.0 & 100.0 & \textbf{58.3} & 50.0 & 66.7 & \textbf{83.3} & 0.17 & 0.17 & 0.17 & 0.17 & 16.00 & \textbf{12.67} & \textbf{10.83} & 11.17 \\
\midrule
Model avg. & 37.5 & \textbf{45.8} & 38.9 & \textbf{47.2} & \textbf{29.2} & 27.8 & 26.4 & \textbf{36.1} & 13.44 & \textbf{12.72} & \textbf{14.22} & 14.61 & 23.97 & \textbf{21.00} & 23.67 & \textbf{21.00} \\
\bottomrule
\end{tabular}}
\end{table*}

\begin{table*}[h]
\centering
\caption{Per-backbone results of external role-play frameworks paired with the same persona-recasting procedure and initial castings. The model-average Final values equal those in \cref{tab:external-frameworks}. Each cell averages 12 episodes; bold marks the better endpoint within each pair.}
\label{tab:external-full}
\scriptsize
\setlength{\tabcolsep}{3pt}
\begin{tabular}{l l | cccc | cccc}
\toprule
& & \multicolumn{4}{c|}{\textbf{NASA Moon} ($L_1\downarrow$)} & \multicolumn{4}{c}{\textbf{Negotiation} (Pareto rate, \%$\uparrow$)} \\
& & \multicolumn{2}{c}{\texttt{Hetero}} & \multicolumn{2}{c|}{\texttt{Random}} & \multicolumn{2}{c}{\texttt{Hetero}} & \multicolumn{2}{c}{\texttt{Random}} \\
\textbf{Framework} & \textbf{Model} & Init. & Final & Init. & Final & Init. & Final & Init. & Final \\
\midrule
SOTOPIA & GPT-4o & \textbf{15.33} & 16.83 & 16.33 & \textbf{15.67} & 0.0 & 0.0 & 0.0 & 0.0 \\
 & GPT-5-mini & \textbf{18.33} & 18.67 & \textbf{16.67} & 18.67 & 0.0 & 0.0 & 0.0 & 0.0 \\
 & Claude Sonnet 4.6 & 12.33 & 12.33 & 12.00 & 12.00 & 0.0 & \textbf{8.3} & 0.0 & 0.0 \\
 & Gemini 3.1 FL & \textbf{15.67} & 16.50 & 20.33 & \textbf{18.33} & \textbf{8.3} & 0.0 & \textbf{16.7} & 0.0 \\
 & Grok 4.20 & 20.67 & \textbf{17.83} & \textbf{16.83} & 19.67 & 0.0 & \textbf{16.7} & \textbf{25.0} & 8.3 \\
 & DeepSeek V4-Flash & 0.00 & 0.00 & 0.17 & 0.17 & 50.0 & \textbf{83.3} & \textbf{58.3} & 33.3 \\
 & \textit{Model avg.} & 13.72 & \textbf{13.69} & \textbf{13.72} & 14.08 & 9.7 & \textbf{18.1} & \textbf{16.7} & 6.9 \\
\midrule
AgentVerse & GPT-4o & 17.00 & 17.00 & 13.83 & 13.83 & \textbf{16.7} & 8.3 & 0.0 & \textbf{25.0} \\
 & GPT-5-mini & \textbf{16.50} & 18.83 & \textbf{15.83} & 18.50 & \textbf{8.3} & 0.0 & 0.0 & \textbf{33.3} \\
 & Claude Sonnet 4.6 & 12.17 & 12.17 & 16.33 & \textbf{12.83} & 0.0 & 0.0 & \textbf{8.3} & 0.0 \\
 & Gemini 3.1 FL & 20.83 & \textbf{18.67} & 21.67 & \textbf{17.33} & \textbf{58.3} & 41.7 & 50.0 & 50.0 \\
 & Grok 4.20 & 19.33 & \textbf{16.83} & 19.00 & \textbf{13.83} & 50.0 & \textbf{58.3} & 16.7 & \textbf{33.3} \\
 & DeepSeek V4-Flash & 0.83 & 0.83 & 1.00 & 1.00 & 8.3 & \textbf{25.0} & 50.0 & \textbf{100.0} \\
 & \textit{Model avg.} & 14.44 & \textbf{14.06} & 14.61 & \textbf{12.89} & \textbf{23.6} & 22.2 & 20.8 & \textbf{40.3} \\
\midrule
Ours & GPT-4o & \textbf{14.33} & 14.83 & 15.33 & \textbf{13.67} & \textbf{75.0} & 50.0 & \textbf{58.3} & 50.0 \\
 & GPT-5-mini & 19.00 & \textbf{18.50} & 21.17 & \textbf{20.83} & 0.0 & \textbf{8.3} & 0.0 & \textbf{8.3} \\
 & Claude Sonnet 4.6 & 8.00 & 8.00 & 10.33 & 10.33 & 33.3 & \textbf{41.7} & \textbf{33.3} & 25.0 \\
 & Gemini 3.1 FL & 17.17 & \textbf{12.00} & 21.17 & \textbf{11.67} & 33.3 & \textbf{58.3} & 41.7 & \textbf{66.7} \\
 & Grok 4.20 & 22.33 & \textbf{16.00} & 18.50 & \textbf{16.83} & 25.0 & \textbf{33.3} & 25.0 & 25.0 \\
 & DeepSeek V4-Flash & 2.83 & 2.83 & 4.17 & 4.17 & 66.7 & \textbf{100.0} & 83.3 & \textbf{91.7} \\
 & \textit{Model avg.} & 13.94 & \textbf{12.03} & 15.11 & \textbf{12.92} & 38.9 & \textbf{48.6} & 40.3 & \textbf{44.4} \\
\bottomrule
\end{tabular}
\end{table*}

\section{Human Annotation Protocol}
\label{app:human-annotation}

We sampled 25 dialogue excerpts by stratified sampling over scenarios, LLM drivers, the two measured dimensions ($R$ and $S$), and the three change directions (increase, unchanged, decrease), and added one attention-check item with an unambiguous answer. Thirteen raters independently labeled each excerpt with the same three-way direction scale used by the judge; all raters passed the attention check. A human-consensus label is assigned by strict majority ($\geq 7$ of 13 raters), and judge agreement is computed only on items with a consensus label. \cref{tab:human-agreement} summarizes the results.

\begin{table}[h]
\centering
\caption{Agreement between the LLM judge and the human majority label.}
\label{tab:human-agreement}
\small
\begin{tabular}{lccc c}
\toprule
\textbf{Dimension} & \textbf{Consensus items} & \textbf{Judge matches} & \textbf{Agreement} & \textbf{Human Fleiss' $\kappa$} \\
\midrule
Relationship ($R$) & 21 & 16 & 76.2\% & 0.288 \\
Emotion ($S$) & 17 & 15 & 88.2\% & 0.184 \\
\bottomrule
\end{tabular}
\end{table}

Inter-human agreement is low, which reflects the inherent ambiguity of reading relational and emotional change from short excerpts rather than a failure of the raters; on the items where humans do converge, the judge follows the majority in most cases. On the relationship dimension, the judge labels a larger share of updates as increasing than the human raters do (about 76\% versus about 40\%). Because every condition in \cref{tab:fidelity-main} and \cref{tab:fidelity-full} is scored by the same judge, such a shared directional tendency affects absolute favorability levels but not the cross-scenario ordering or the cross-model comparisons on which our conclusions rest.

\section{Full Social-Dynamics Fidelity Results}
\label{app:fidelity-full}

\cref{tab:fidelity-full} reports all 30 scenario--model conditions underlying \cref{tab:fidelity-main}, as mean $\pm$ standard deviation over five complete nine-round episodes per condition. Panel~(a) of \cref{tab:fidelity-main} is the equally weighted mean of the six model rows within each scenario, and panel~(b) corresponds to the Tavern rows. The last column reports the Kruskal--Wallis test of final favorability across the six models within each scenario (df $=5$).

\begin{table*}[h]
\centering
\caption{Full social-dynamics fidelity results (mean $\pm$ s.d., $n{=}5$ episodes per condition). Metric definitions follow \cref{app:metric-defs}. KW: Kruskal--Wallis test of $\mathrm{Fav}^{T}$ across models within each scenario.}
\label{tab:fidelity-full}
\scriptsize
\setlength{\tabcolsep}{3pt}
\renewcommand{\arraystretch}{1.08}
\resizebox{\textwidth}{!}{%
\begin{tabular}{l l cccccc c}
\toprule
\textbf{Scenario} & \textbf{Model} & $\mathrm{Fav}^{T}$ & $S^{T}$ & Recip. Corr. & E-I index & Balance (\%) & Repair (\%) & KW \\
\midrule
\multirow{6}{*}{Hospital ethics review} & GPT-5-mini & $0.82 \pm 0.03$ & $0.02 \pm 0.02$ & $0.36 \pm 0.26$ & $-0.10 \pm 0.02$ & $100.0 \pm 0.0$ & $40.0 \pm 20.0$ &  \\
 & Claude Sonnet 4.6 & $0.86 \pm 0.06$ & $0.18 \pm 0.07$ & $0.28 \pm 0.17$ & $-0.08 \pm 0.05$ & $100.0 \pm 0.0$ & $73.3 \pm 38.9$ &  \\
 & DeepSeek V4-Flash & $0.79 \pm 0.04$ & $0.24 \pm 0.06$ & $0.49 \pm 0.21$ & $-0.17 \pm 0.04$ & $100.0 \pm 0.0$ & $80.0 \pm 40.0$ & $H{=}23.36$ \\
 & Gemini 3.1 FL & $0.74 \pm 0.03$ & $0.05 \pm 0.06$ & $0.76 \pm 0.12$ & $-0.14 \pm 0.08$ & $100.0 \pm 0.0$ & $58.0 \pm 15.7$ & $p{=}0.0003$ \\
 & Grok 4.20 & $0.54 \pm 0.04$ & $0.04 \pm 0.02$ & $-0.08 \pm 0.31$ & $0.05 \pm 0.08$ & $100.0 \pm 0.0$ & $39.7 \pm 9.7$ &  \\
 & GPT-4o & $0.74 \pm 0.02$ & $0.11 \pm 0.05$ & $0.32 \pm 0.31$ & $-0.15 \pm 0.04$ & $100.0 \pm 0.0$ & $60.0 \pm 49.0$ &  \\
\midrule
\multirow{6}{*}{Tavern inheritance dispute} & GPT-5-mini & $0.75 \pm 0.03$ & $0.23 \pm 0.05$ & $0.67 \pm 0.28$ & $-0.22 \pm 0.07$ & $100.0 \pm 0.0$ & $60.0 \pm 49.0$ &  \\
 & Claude Sonnet 4.6 & $0.57 \pm 0.07$ & $0.42 \pm 0.11$ & $0.93 \pm 0.02$ & $-0.53 \pm 0.13$ & $100.0 \pm 0.0$ & $80.0 \pm 24.5$ &  \\
 & DeepSeek V4-Flash & $0.57 \pm 0.05$ & $0.32 \pm 0.04$ & $0.64 \pm 0.11$ & $-0.52 \pm 0.05$ & $94.0 \pm 12.0$ & $53.3 \pm 24.5$ & $H{=}19.66$ \\
 & Gemini 3.1 FL & $0.51 \pm 0.10$ & $0.13 \pm 0.07$ & $0.69 \pm 0.12$ & $-0.67 \pm 0.12$ & $80.0 \pm 16.7$ & $48.8 \pm 25.7$ & $p{=}0.0014$ \\
 & Grok 4.20 & $0.47 \pm 0.05$ & $0.24 \pm 0.10$ & $0.61 \pm 0.15$ & $-0.63 \pm 0.12$ & $100.0 \pm 0.0$ & $56.7 \pm 38.9$ &  \\
 & GPT-4o & $0.45 \pm 0.02$ & $0.46 \pm 0.13$ & $0.92 \pm 0.02$ & $-0.72 \pm 0.06$ & $100.0 \pm 0.0$ & $10.0 \pm 20.0$ &  \\
\midrule
\multirow{6}{*}{Adventurer-party quest} & GPT-5-mini & $0.54 \pm 0.04$ & $0.22 \pm 0.06$ & $0.77 \pm 0.14$ & $-0.55 \pm 0.11$ & $88.0 \pm 14.7$ & $30.5 \pm 17.2$ &  \\
 & Claude Sonnet 4.6 & $0.61 \pm 0.04$ & $0.27 \pm 0.06$ & $0.78 \pm 0.09$ & $-0.46 \pm 0.11$ & $100.0 \pm 0.0$ & $80.8 \pm 10.5$ &  \\
 & DeepSeek V4-Flash & $0.53 \pm 0.03$ & $0.29 \pm 0.02$ & $0.81 \pm 0.11$ & $-0.54 \pm 0.12$ & $80.0 \pm 16.7$ & $40.0 \pm 28.9$ & $H{=}22.39$ \\
 & Gemini 3.1 FL & $0.40 \pm 0.06$ & $0.15 \pm 0.06$ & $0.70 \pm 0.12$ & $-0.70 \pm 0.15$ & $60.0 \pm 15.5$ & $46.3 \pm 9.7$ & $p{=}0.0004$ \\
 & Grok 4.20 & $0.40 \pm 0.06$ & $0.37 \pm 0.08$ & $0.78 \pm 0.06$ & $-0.77 \pm 0.10$ & $72.0 \pm 16.0$ & $15.2 \pm 11.0$ &  \\
 & GPT-4o & $0.48 \pm 0.02$ & $0.23 \pm 0.04$ & $0.87 \pm 0.03$ & $-0.77 \pm 0.07$ & $72.0 \pm 14.7$ & $22.4 \pm 19.6$ &  \\
\midrule
\multirow{6}{*}{Solar-farm negotiation} & GPT-5-mini & $0.52 \pm 0.06$ & $0.19 \pm 0.05$ & $0.86 \pm 0.07$ & $-0.43 \pm 0.17$ & $94.0 \pm 12.0$ & $27.0 \pm 6.0$ &  \\
 & Claude Sonnet 4.6 & $0.54 \pm 0.03$ & $0.20 \pm 0.03$ & $0.84 \pm 0.07$ & $-0.53 \pm 0.08$ & $88.0 \pm 14.7$ & $18.0 \pm 10.7$ &  \\
 & DeepSeek V4-Flash & $0.47 \pm 0.03$ & $0.12 \pm 0.02$ & $0.94 \pm 0.02$ & $-0.77 \pm 0.12$ & $60.0 \pm 0.0$ & $19.8 \pm 5.9$ & $H{=}22.56$ \\
 & Gemini 3.1 FL & $0.29 \pm 0.06$ & $0.11 \pm 0.04$ & $0.98 \pm 0.02$ & $-0.96 \pm 0.22$ & $92.0 \pm 16.0$ & $6.7 \pm 13.3$ & $p{=}0.0004$ \\
 & Grok 4.20 & $0.39 \pm 0.04$ & $0.10 \pm 0.06$ & $0.92 \pm 0.02$ & $-1.04 \pm 0.10$ & $72.0 \pm 14.7$ & $24.2 \pm 17.6$ &  \\
 & GPT-4o & $0.39 \pm 0.03$ & $0.17 \pm 0.09$ & $0.93 \pm 0.02$ & $-1.03 \pm 0.13$ & $66.0 \pm 4.9$ & $12.9 \pm 12.4$ &  \\
\midrule
\multirow{6}{*}{Demolition standoff} & GPT-5-mini & $0.28 \pm 0.04$ & $0.29 \pm 0.05$ & $0.24 \pm 0.18$ & $-2.01 \pm 0.44$ & $94.0 \pm 12.0$ & $26.1 \pm 3.8$ &  \\
 & Claude Sonnet 4.6 & $0.27 \pm 0.05$ & $0.23 \pm 0.05$ & $0.80 \pm 0.04$ & $-2.44 \pm 0.56$ & $54.0 \pm 4.9$ & $16.8 \pm 6.4$ &  \\
 & DeepSeek V4-Flash & $0.28 \pm 0.03$ & $0.23 \pm 0.06$ & $0.74 \pm 0.05$ & $-2.33 \pm 0.50$ & $94.0 \pm 12.0$ & $41.3 \pm 7.5$ & $H{=}18.73$ \\
 & Gemini 3.1 FL & $0.08 \pm 0.03$ & $0.14 \pm 0.04$ & $0.80 \pm 0.02$ & $-1.20 \pm 18.13$ & $64.0 \pm 4.9$ & $11.8 \pm 6.6$ & $p{=}0.0022$ \\
 & Grok 4.20 & $0.19 \pm 0.01$ & $0.25 \pm 0.03$ & $0.71 \pm 0.02$ & $-3.43 \pm 0.37$ & $68.0 \pm 17.2$ & $14.7 \pm 14.6$ &  \\
 & GPT-4o & $0.28 \pm 0.02$ & $0.27 \pm 0.06$ & $0.78 \pm 0.08$ & $-2.08 \pm 0.28$ & $56.0 \pm 4.9$ & $2.9 \pm 5.7$ &  \\
\bottomrule
\end{tabular}}
\end{table*}

\begin{figure*}[h]
\centering
\includegraphics[width=\linewidth]{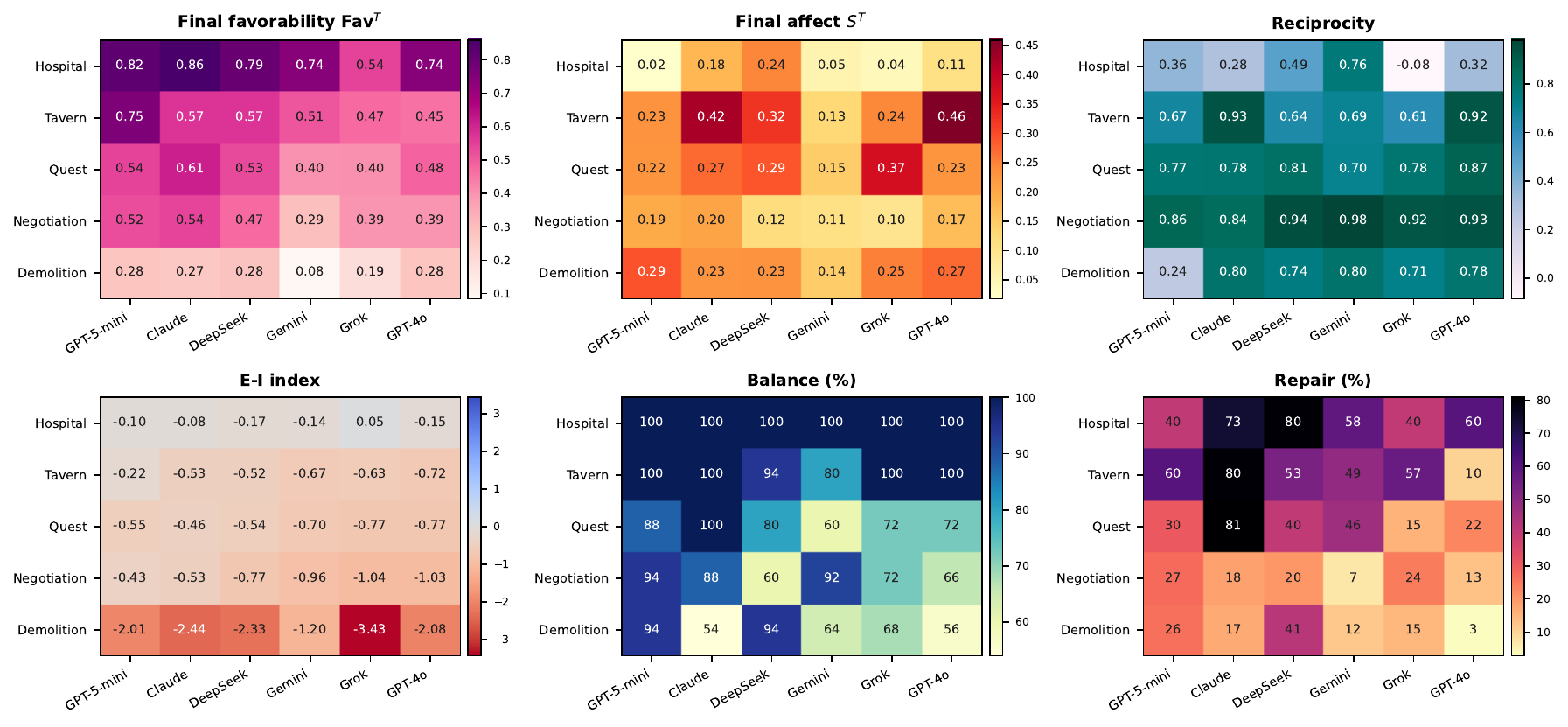}
\caption{Scenario $\times$ model means of the six fidelity metrics ($n{=}5$ episodes per cell).}
\label{fig:fidelity-heatmaps}
\end{figure*}

\begin{figure*}[h]
\centering
\includegraphics[width=\linewidth]{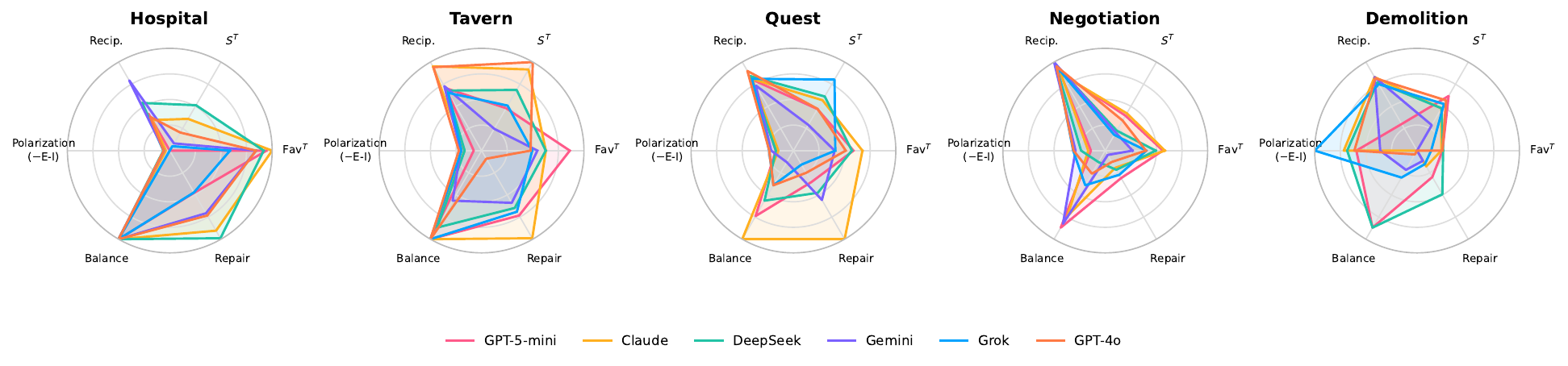}
\caption{Social-dynamics profiles of the six models in each scenario. Axes are min--max normalized across all 30 conditions; polarization is $-$E-I.}
\label{fig:fidelity-radar}
\end{figure*}

\cref{fig:fidelity-heatmaps} visualizes the same 30 conditions metric by metric, and \cref{fig:fidelity-radar} overlays the six models within each scenario. Each radar axis is min--max normalized over all 30 conditions, and polarization is plotted as $-$E-I so that larger values indicate stronger factional separation. The scenario ordering is visible as a shrinking favorability and repair profile from Hospital to Demolition, while model profiles remain distinct within every scenario.

\section{Dynamic Trajectories}
\label{app:trajectories}

\begin{figure*}[h]
\centering
\includegraphics[width=\linewidth]{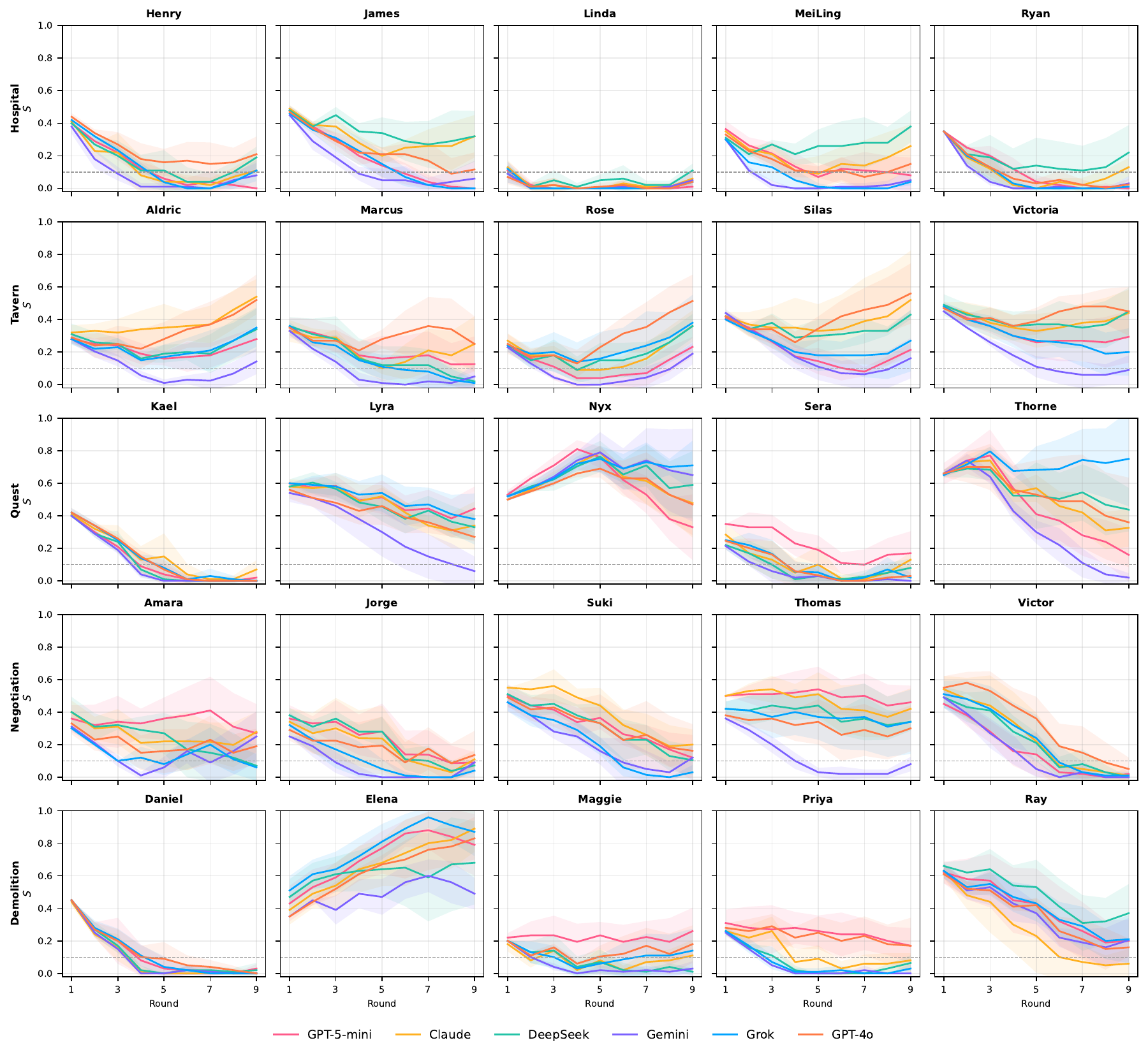}
\caption{Emotional-state trajectories of every agent across nine rounds. Lines show the mean over five episodes and shaded regions one standard deviation; the dashed line marks $S=0.1$.}
\label{fig:s-trajectory-all}
\end{figure*}

\begin{figure}[h]
\centering
\includegraphics[width=0.75\linewidth]{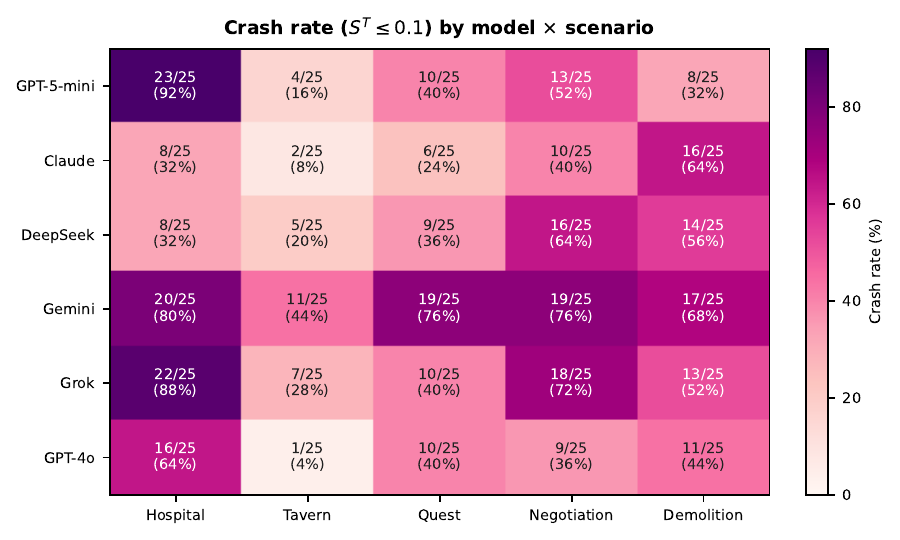}
\caption{Crash rate: proportion of agent-episode final states with $S^{T}\leq 0.1$ (25 agent-episodes per cell).}
\label{fig:crash-heatmap-all}
\end{figure}

Terminal metrics hide how emotional states and relationships evolve. The figures below are computed from the same 150 episodes as \cref{tab:fidelity-full} (five complete nine-round episodes per scenario--model condition). \cref{fig:s-trajectory-all} shows per-agent emotional trajectories, \cref{fig:fav-trajectory-main} shows team-level mean favorability over all directed edges, and \cref{fig:crash-heatmap-all} reports the proportion of agent-episode final states with $S^{T}\leq 0.1$.

\end{document}